\documentclass[journal]{IEEEtran}
\ifCLASSINFOpdf
\else
\fi
\usepackage{amsmath}
\usepackage{graphicx,dblfloatfix} 

\usepackage{float} 
\usepackage{wrapfig} 
\usepackage{dblfloatfix}    
\usepackage{amsmath}
\usepackage{subcaption}
\usepackage{rotating}
\usepackage[font={small}]{caption}
\usepackage{algorithm}
\usepackage[noend]{algpseudocode}

\usepackage{romannum}
\usepackage[table,xcdraw]{xcolor}
\usepackage{comment}
\graphicspath{{Pictures/}}
\usepackage{amsmath,mathtools}
\usepackage{amsmath}
\usepackage{tikz}
\usepackage{mathdots}
\usepackage{yhmath}
\usepackage{cancel}
\usepackage{color}
\usepackage{siunitx}
\usepackage{array}
\usepackage{multirow}
\usepackage{amssymb}
\usepackage{gensymb}
\usepackage{tabularx}
\usepackage{booktabs}
\usetikzlibrary{fadings}
\usetikzlibrary{patterns}
\usetikzlibrary{shadows.blur}
\usetikzlibrary{shapes}
\usepackage[normalem]{ulem}

\usepackage{graphics} 
\makeatletter
\def\endthebibliography{%
	\def\@noitemerr{\@latex@warning{Empty `thebibliography' environment}}%
	\endlist
}
\makeatother

\begin{document}
%

\title{Image-based Intraluminal Contact Force Monitoring in Robotic Vascular Navigation}

%
%
%
\author{Masoud~Razban,~\IEEEmembership{Member,~IEEE,}
	Javad~Dargahi,~\IEEEmembership{Member,~IEEE,}
	Benoit~Boulet,~\IEEEmembership{Senior~Member,~IEEE}
	\thanks{This work was supported by Natural	Sciences and Engineering Research Council of Canada (NSERC), CREATE Innovation at the Cutting Edge grant. }
	\thanks{Masoud Razban is with Centre for Intelligent Machines (CIM) at McGill University and Robotic Surgery Lab at Concordia University, Montreal, QC, CA H3A 2A7
		{\tt\footnotesize (masoud.razban@mail.mcgill.ca)}}
	\thanks{Dr. Javad Dargahi is with Robotic Surgery Lab in Mechanical Engineering Department, Concordia University, Montreal, QC, CA, H3G 1M8
		{\tt\footnotesize (dargahi@encs.concordia.ca)}}
	\thanks{Dr. Benoit Boulet is with Centre for Intelligent Machines (CIM) at McGill University, Montreal, QC, CA, H3A 2A7	{\tt\footnotesize(benoit.boulet@mcgill.ca) }}}

%

\markboth{
}%
{Shell \MakeLowercase{\textit{et al.}}: Bare Demo of IEEEtran.cls for IEEE Journals}
%



\maketitle

\begin{abstract}
Embolization, stroke, ischaemic lesion, and perforation remain significant concerns in endovascular interventions. Intravascular sensing of tool interaction with the arteries is advantageous to minimize such complications and enhance navigation safety. Intraluminal information is currently limited due to the lack of intravascular contact sensing technologies. We present monitoring of the intraluminal tool interaction with the arterial wall using an image-based estimation approach within vascular robotic navigation. The proposed image-based method employs continuous finite element simulation of the tool using imaging data to estimate multi point forces along tool-vessel wall interaction. We implemented imaging algorithms to detect and track contacts, and compute pose measurements. The model is constructed based on the nonlinear beam element and flexural rigidity profile over the tool length. During remote cannulation of aortic arteries, intraluminal monitoring achieved tracking local contact forces, building a contour map of force on the arterial wall and estimating tool structural stress. Results suggest that high risk intraluminal forces may happen even with low insertion force. The presented online monitoring system delivers insight into the intraluminal behavior of endovascular tools and is well-suited for intraoperative visual guidance for the clinician, robotic control of vascular procedures and research on interventional device design.
\end{abstract}


%
\IEEEpeerreviewmaketitle

\section{Introduction}
\IEEEPARstart{E}{ndovascular} interventions are leading treatments and diagnoses for cardiovascular disease. Despite improvements in tools and techniques, intraprocedural risks of embolization, ischaemic brain lesions, and stroke in percutaneous procedures are still high, especially in carotid artery stenting \cite{kakisis2012european,ederle2009randomized}.
Studies reported that 50\% of the cases after carotid stenting had a new ischaemic lesion on diffusion-weighted imaging (DWI) of the post-treatment scan due to embolism. \cite{gensicke2013characteristics}. The high number of microemboli, which has been reported during navigation of catheters and guidewires, highlights the importance of tool-tissue interaction \cite{ackerstaff2003transcranial, Kim1732}. Studies have further pointed out other complications, perforation, thrombosis and dissection, as a result of catheter/guidewire interactions with the arterial wall \cite{hausegger2001complications}. In the case of stenosis treatment, excess insertion force to cross occlusion could raise the likelihood of perforating \cite{singh2009endovascular}. Limited motion of under-actuated conventional catheter and guidewire can further add to the risk, especially in the case of a diseased and torturous vessel. These findings suggest that the intraluminal interaction contact force (CF) is one of the determinants of patient safety and procedure efficiency. Monitoring force information intraoperatively has the potential to enhance navigation safety and efficiency. Practical applications are threefold: 
\begin{itemize}
	\item Integrating CF data into intraoperative visual guidance for clinicians to bring safe catheter/guidewire manipulation. Intraluminal insight could minimize the risk of stroke, embolism, vessel perforation, or dissection. It could further limit human mishandling, especially in the case of novices, with implications to improve training for complex procedures.
	\item The development of an automated robotic surgery system that maintains CF in a safe range and improves smoothness using realtime intraluminal data. Vascular robotic technology demonstrated definite advantages in catheter controllability, stability, precision, and lower radiation for clinicians \cite{antoniou2011clinical,smilowitz2012robotic, carrozza2012robotic, riga2013robot, bao2018operation}. With the advent of artificial intelligence, a safe and autonomous form of robotic surgery can be introduced \cite{chen2020deep,bao2018operation}.
	\item Intraluminal force information, along with catheter deflection data can also be used for research on design of interventional tools aimed to minimize vessel injuries and maximize maneuverability, torquability and deliverability. It would be beneficial to study the intraluminal performance of interventional devices and optimize their design.
\end{itemize}

Intraluminal interaction information is limited due to the lack of force sensing elements along the endovascular tools or other remote sensing technologies. Tool-tissue contact force sensing remained limited to the tip single point force measurement. In electrophysiology, proper contact of the catheter tip and contact force are essential due to the chance of over-burn in case of excessive CF or inferior lesion quality in weak contact \cite{kuck2012novel}. Studies proposed ablation catheters with tip force sensors \cite{polygerinos2013triaxial, strandman1997production, kesner2011position} and some are available commercially \cite{Biosense56:online, TactiCat78:online}. Instrumented cardiovascular tools remain limited to tip-mounted force sensing in which the complexity of integrating sensing elements, cost, maneuverability, and interference with nearby devices are some of the associated challenges. The instrumentation of guidewires is more challenging due to their smaller diameter and complex structure. Indirect force estimation at the tip of ablation catheter has worked based on modeling and shape analysis, e.g., kinematic model, cosserat rod, and piecewise planar elastica: \cite{khoshnam2015modeling, Shahir,back2015catheter, ganji2009catheter, hasanzadeh2014efficient} has reported promising results. On the other hand, proximal insertion force measurement and methods to control it have been used in robotic catheterization \cite{jayender2009robot}, even though this force does not represent local vessel interaction forces. A quantitative analysis of the total contact force between the instrument and cardiovascular phantoms has been established in robotic navigation \cite{rafii2016reducing} as well and used as an assessment metric \cite{rafii2017objective,chi2018learning} of navigation, whereas the relation between the total exerted force and local intraluminal CF is not entirely clear.

Monitoring intraluminal vessel contact forces throughout the length of the tool is a gap in the literature. In previous work \cite{razban2018sensor}, we presented and validated a sensor-less approach to estimate multi-point load forces at the side of interventional tools. This study aims to develop the proposed approach to intraluminal interaction force monitoring within navigation of an anthropomorphic cardiovascular phantom and to employ a teleoperated robotic platform. The FEM formulation has been improved to consider shear deformation based on Timoshenko beam theory. Image segmentation algorithms have been updated for intravascular contact tracking and pose measurements within the phantom. This study also presents a quantitative analysis to investigate the resultant force exerted on the vasculature compared with intraluminal contact forces (ICF), to highlight the potential need for local tool-tissue monitoring. A two-degree freedom\,(DOF) insertion robot is designed and fabricated based on the methods in conventional manual navigation which allow simultaneous and independent control of insertion and rotation. The 6 DOF force/torque sensor is coupled to the aortic arch phantom to measure the resultant force exerted on the vasculature. catheter/guidewire, as a fixed model parameter input, is measured through a sequential bending test over its length. The nonlinear finite element model is set to simulate the tool navigation using real-time image-based pose measurements. Solving the inverse FEM model estimates the contact forces and catheter's deflected shape. We present the contour map of intraluminal CF on vessel boundaries and the tracking of multiple local ICF. An intravascular stress analysis of the tool is also conducted via the proposed method. In the manuscript, "catheter" refers to any general type of non-steerable catheters and guidewires for conventional endovascular intervention.  
\section{Methods and Algorithm}
\subsection{Contact force estimation}
The accuracy of the proposed sensor-less force estimation concept was shown in the previous work \cite{razban2018sensor} through a direct force measurement setup, i.e., single contact point phantoms mounted on a 6-DOF F/T sensor. An estimation accuracy of over 87\% was achieved in agreement with force sensor measurements \cite{razban2018sensor}. The proposed method is based on continuous simulation of catheter deflection as a beam using data from real-time imaging. Image processing segments and tracks the deflected catheter shape and locates the interactions with the internal wall of the vasculature. Using image processing data and prior knowledge about the intrinsic shape of the catheter, an FEM model is built and solved. Deflections induced by the vessel wall at the contact points set the boundary conditions of the model, and reaction forces computed from these boundary conditions return the contact forces. Deflections are fed to the model continuously to simulate catheter manipulation. The FEM model is based on the Timoshenko beam element, considering nonlinear large deformation. The model assumes that the catheter structure is a uniform round beam with an equivalent bending modulus. 
\begin{figure}[tb]
	\centering
	\includegraphics[width=0.42\textwidth]{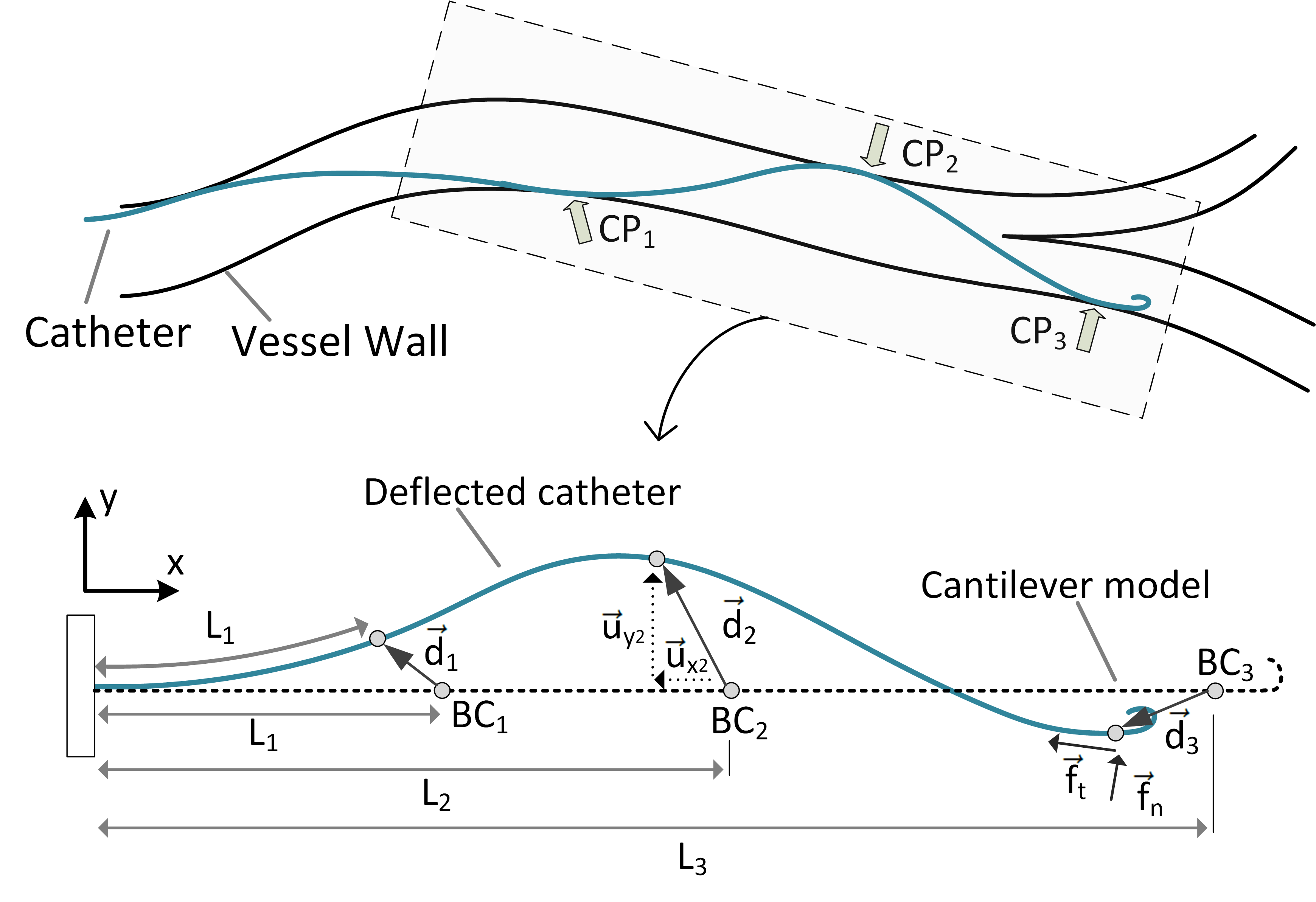}	
	\caption{Force estimation concept showing deflected catheter and constructed cantilever model. $ CP_i $s are the contact points with length of $ L_i $ on the deflected shape and $ BC_i $s are the associated boundary condition on the model.}
	\label{forceestimator}
\end{figure}

Let us describe a schematic view of a catheter-vessel interaction as shown in Fig.~\ref{forceestimator}. A section of the catheter beginning at any desired point and ended at the tip is modeled as a cantilever, which has all contact points included (window shown in Fig.~\ref{forceestimator}). The model is created in the local coordinate of the catheter, having the $ \vec{x} $ axis tangent to its base. This is to compute deflection from the catheter base perspective and isolate it from the global coordinates. The cantilever model has an intrinsic catheter body of length equal to the deflected shape in the window, $ L_{cantilever}=L_{tip} $. The model length varies as the catheter advances or retracts during navigation. The initial positions of contacts points on the model have equal lengths as on the deflected catheter. For instance, the position of contact point $i$ is $ CP_i $, which has length $ L_i $ along the deflected shape, and the corresponding point with the same length on the cantilever model would be the boundary condition $ BC_i $, i.e., it is the contact point before being deflected. Contact deflection $ \vec{d_i} $ is the displacement between the deflected contact point position, $ P_{CPi} $, and its rest position on the model, $ P_{BCi} $ (see (\ref{eq.1}) and (\ref{eq.2}) ). Deflections, $\psi(d)=\{\vec{d_i}\}$, are applied to boundary conditions $BC=\{BC_i\} $, respectively. By solving the inverse FEM model, the catheter is being deflected as it is interacting with the vessel, which simulates the current state. The FEM model computes the reaction forces, $ \bar{CF}= \{\bar{F_i}\} $, at boundary conditions as estimates of the contact forces on the vasculature. Even though the magnitude of contact forces matters for safety, CF can be broken into components of normal ($\vec{f_n}$) and tangential/frictional forces ($\vec{f_t}$) if such information is needed. As the catheter navigates through the vasculature, its deflected shape, the position of contact points, and the amount of deflection change. The cantilever FEM model parameters and computed contact forces are updated in real-time accordingly.
\begin{equation}
\label{eq.1}
\vec{d}=\vec{u}_x+\vec{u}_y=\begin{bmatrix}u_x & u_y\end{bmatrix}
\end{equation}
\begin{equation}
\label{eq.2}
{\footnotesize \psi(d)=\begin{bmatrix} \vec {d_{1}}\\\vdots\\ \vec{d_{i}}\end{bmatrix}=\begin{bmatrix} u_{x1}&u_{y1}\\\vdots & \vdots\\u_{xi}&u_{yi}\end{bmatrix}= \\
\begin{bmatrix} x_{CP_{1}}-x_{BC_{1}}&y_{CP_{1}}-y_{BC_{1}} \\ \vdots & \vdots \\x_{CP_{i}}-x_{BC_{i}}&y_{CP_{i}}-y_{BC_{i}} \end{bmatrix}}
\end{equation}
In our experiments, a planar projection model is adapted to demonstrate practical feasibility as a proof of concept for ICF monitoring where the distal section of the tool almost lays within a plane. Thus, in this particular test case, the effect of out-of-plane deflection is considered minor compared to the large in-plane deflections. This might not apply to other endovascular procedures requiring a 3D implementation.

\subsection{Image segmentation and contact tracking}
Image segmentation obtains data required for the model, including lengths from catheter base to every contact point and tip, positions and deflections of contact points in the local coordinates of the catheter (see Fig.~\ref{forceestimator}). The OpenCV (Open Source Computer Vision) library is used for image processing programmed in C++. Images are continuously sampled through an RGB camera with 1920*1080 pixels at 30 frame per second (FPS). The RGB image is converted to grayscale to be similar to X-ray fluoroscopic images. Medical imaging and visualization systems (X-ray fluoroscopy, CT and MRI) enable the detection and tracking of the shape of the catheter and vessel boundaries in cardiovascular interventions. In this study, we detect the vessel using the contrast between the vascular phantom and the background, similar to injecting contrast agent to visualize the lumen in X-ray fluoroscopy. We assume there is no extreme movement or change in vessels during the procedure; however, an online calibration technique can be used to match the vessel image in case of any movement like breathing in a clinical case.

\begin{figure}[tb]
	\centering
	\includegraphics[width=0.45\textwidth,trim={0 0.2cm 0 0.2cm},clip]{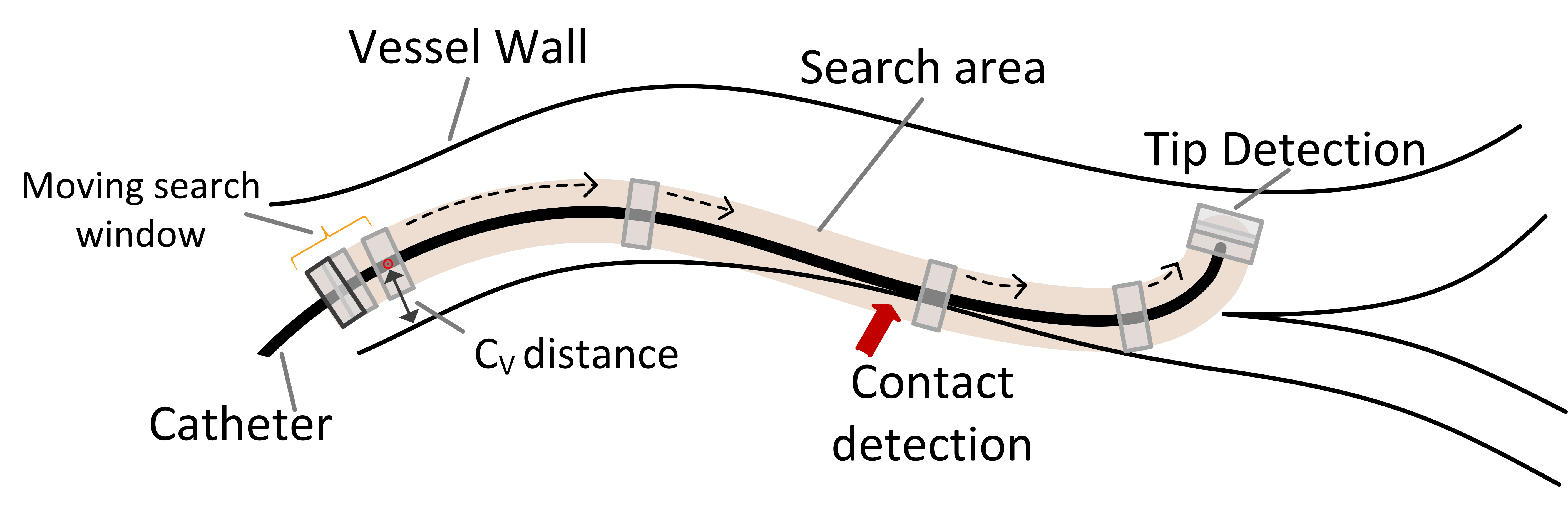}
	\caption{Moving search window sweeps catheter along its length to extract centerline, and detect contacts and tip.}
	\label{search}
\end{figure}

\begin{figure*}[tb]
	\centering
	\begin{subfigure}{0.21\textwidth}
		\includegraphics[width=\textwidth]{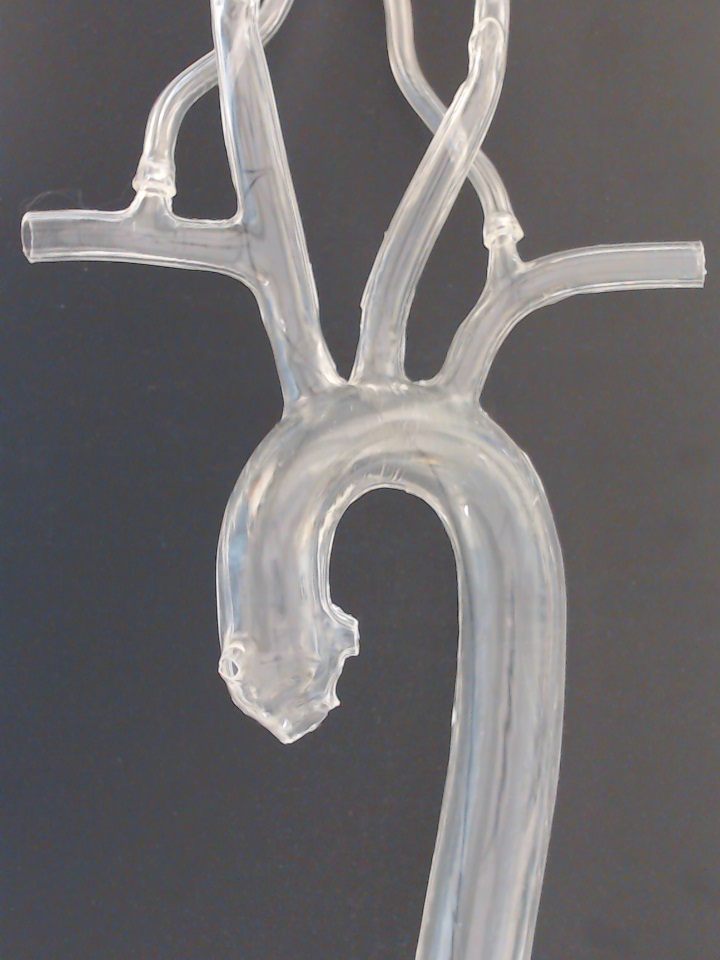}
		\caption{}
	\end{subfigure}%
\hspace{1mm}
	\begin{subfigure}{0.21\textwidth}
		\includegraphics[width=\textwidth]{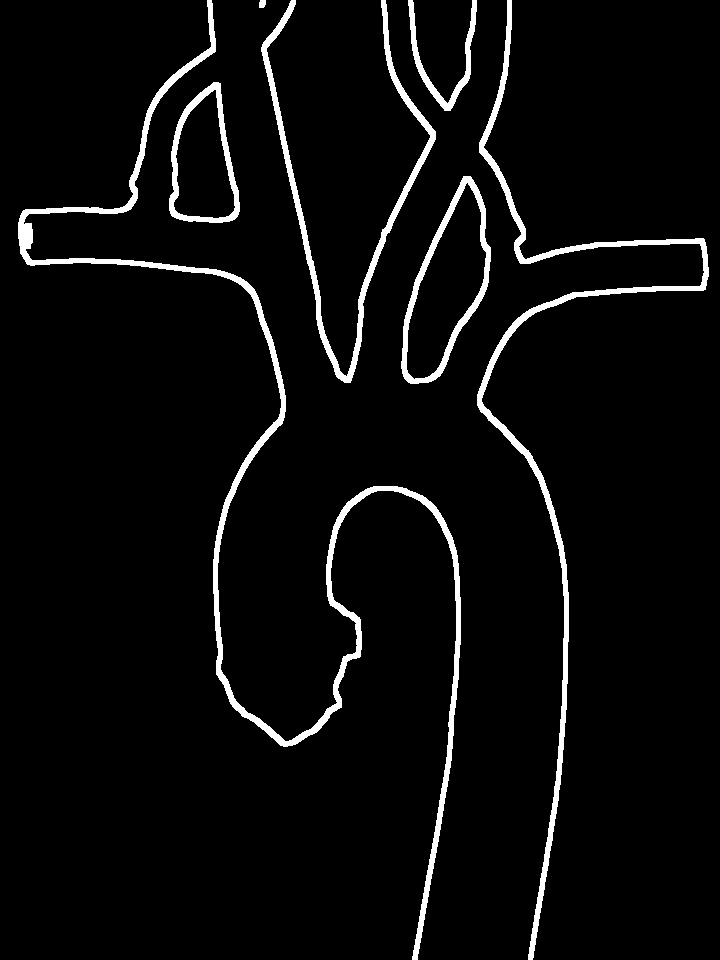}
		\caption{}
	\end{subfigure}
\hspace{0.1mm}
	\begin{subfigure}{0.21\textwidth}
		\includegraphics[width=\textwidth]{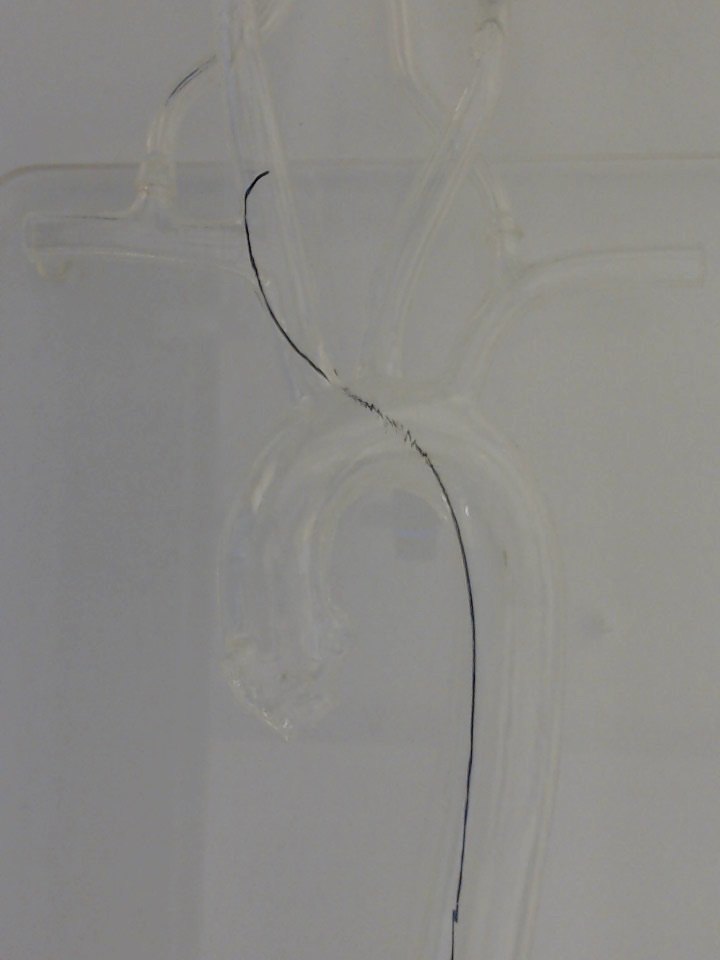}
		\caption{}
	\end{subfigure}%
\hspace{1mm}
	\begin{subfigure}{0.21\textwidth}
		\includegraphics[width=\textwidth]{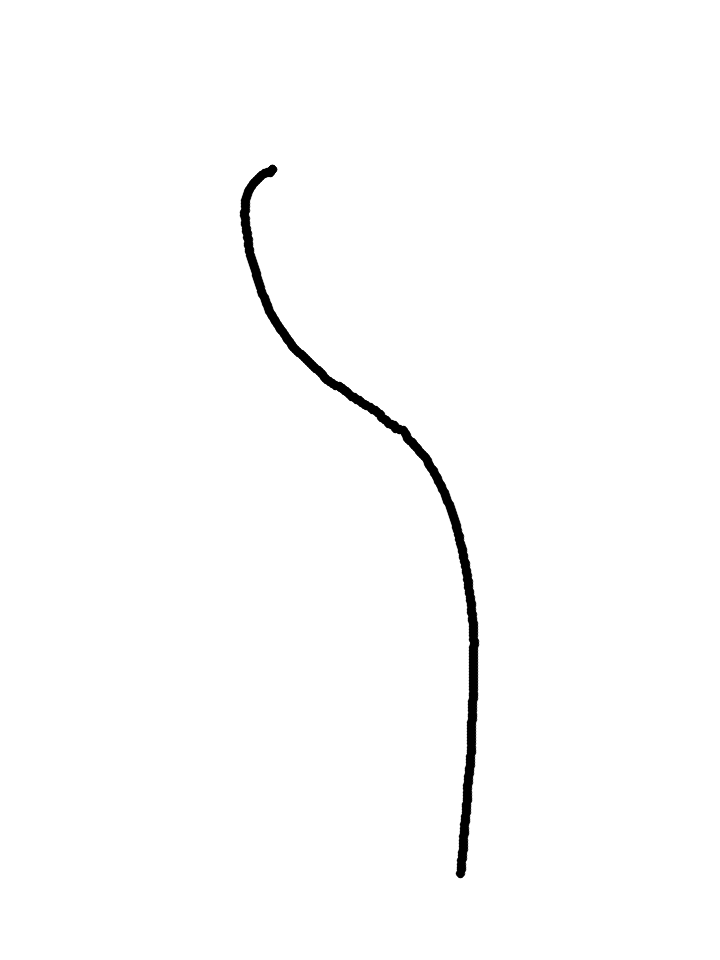}
		\caption{}
	\end{subfigure}
	\begin{subfigure}{0.21\textwidth}
		\includegraphics[width=\textwidth]{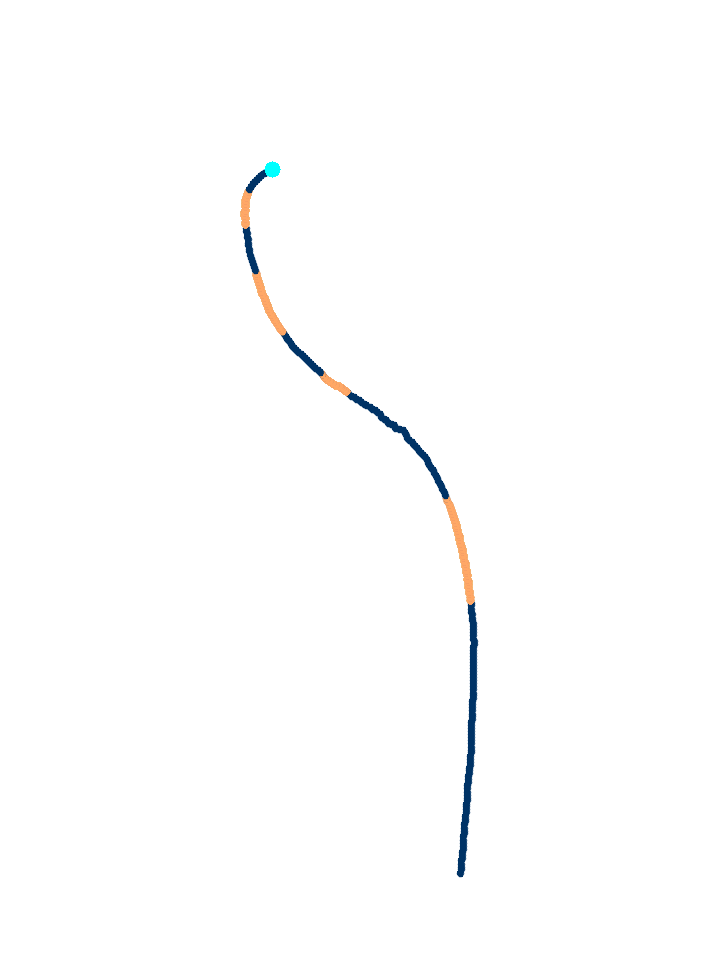}
		\caption{}
	\end{subfigure}
\hspace{1mm}
	\begin{subfigure}{0.21\textwidth}
		\includegraphics[width=\textwidth]{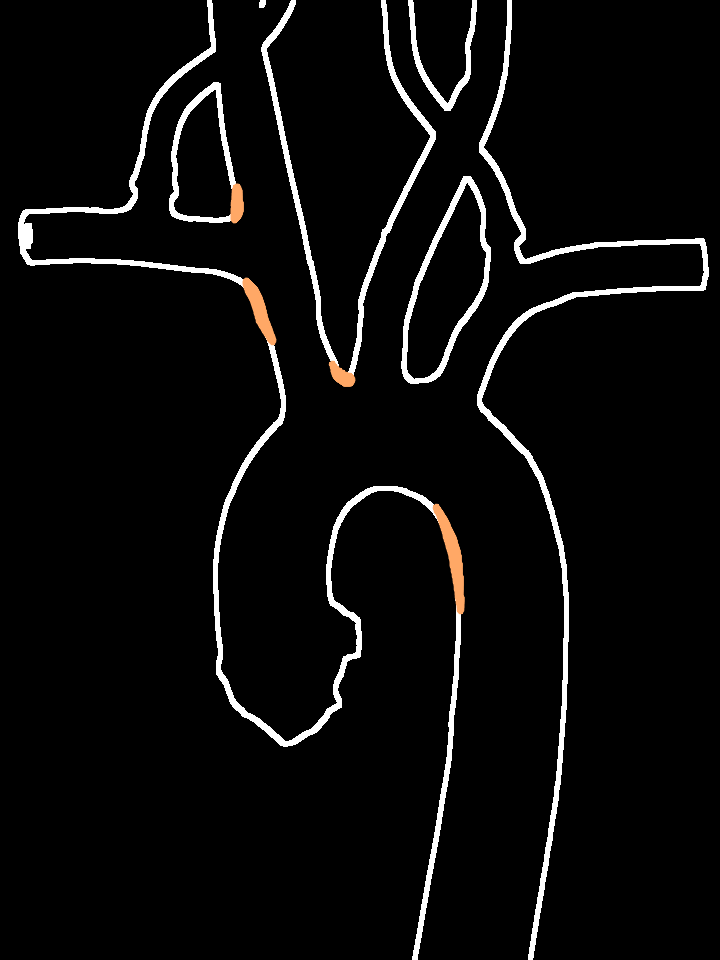}
		\caption{}
	\end{subfigure}
\hspace{0.1mm}
	\begin{subfigure}{0.21\textwidth}
		\includegraphics[width=\textwidth]{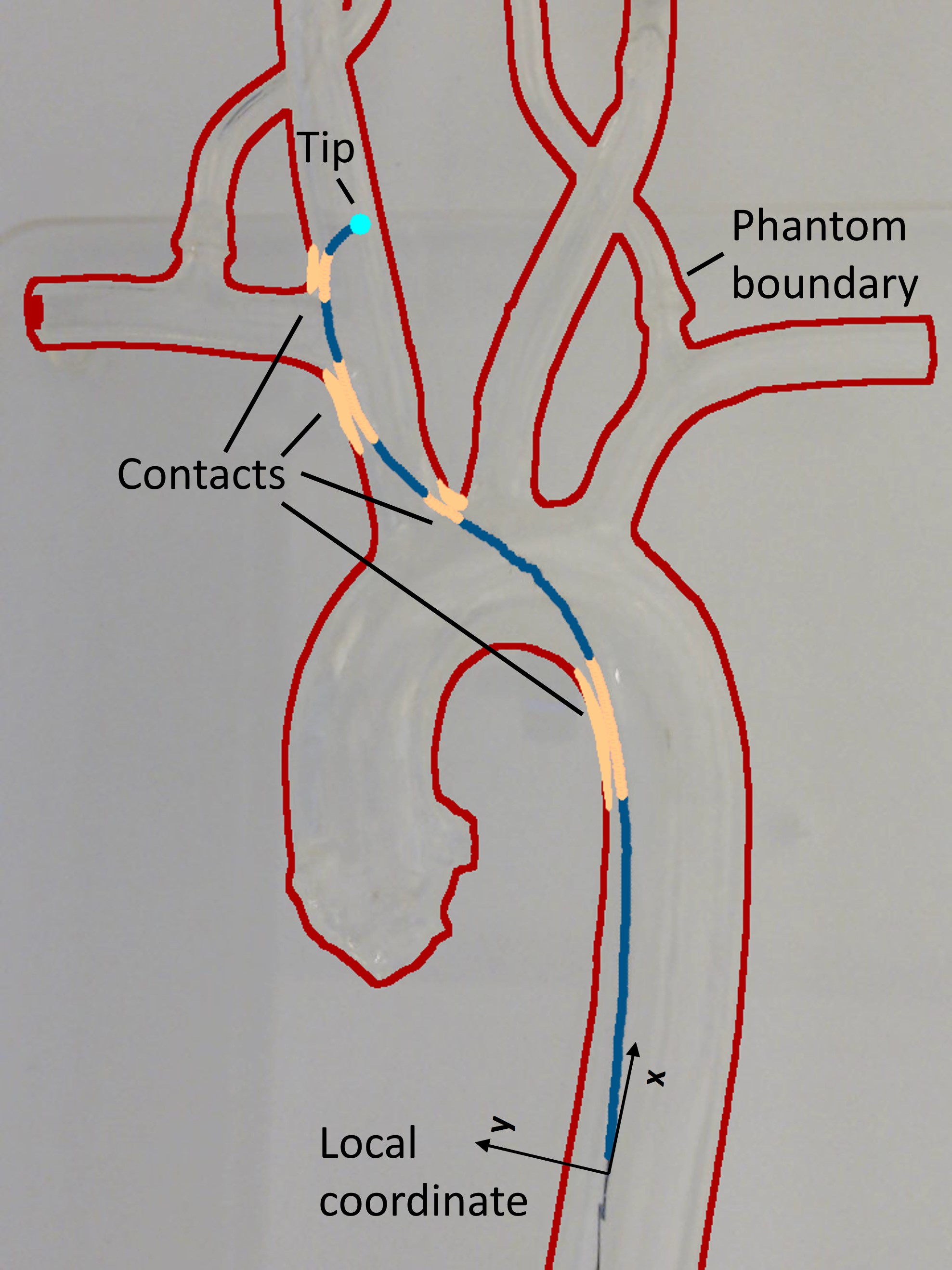}
		\caption{}
	\end{subfigure}	
\hspace{1mm}
	\begin{subfigure}{0.21\textwidth}
		\includegraphics[width=\textwidth]{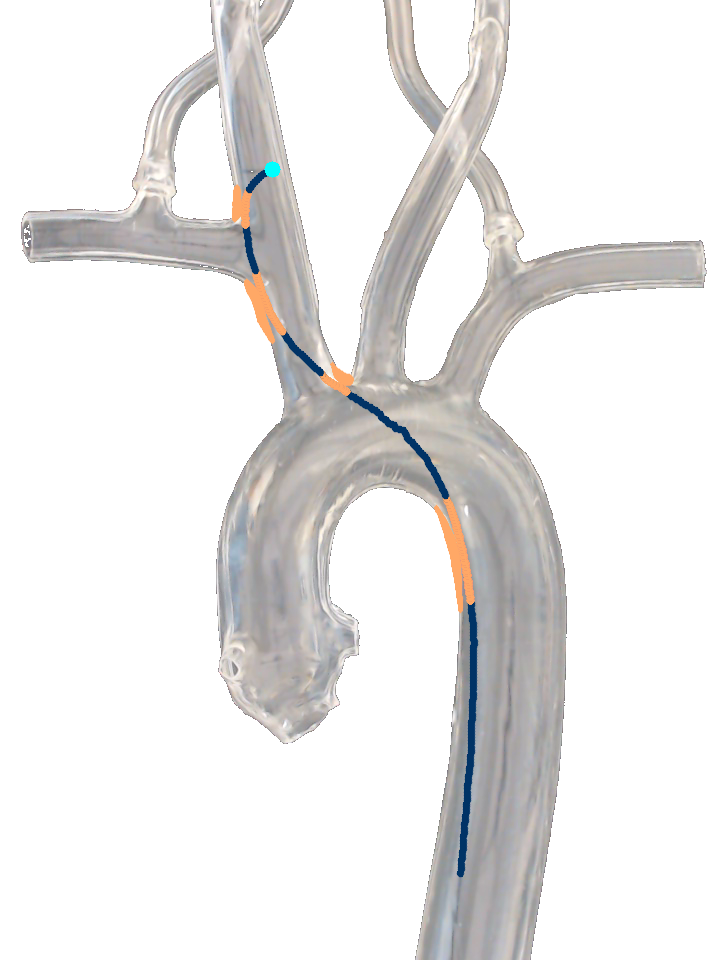}
		\caption{}
	\end{subfigure}	
	\captionsetup{margin=0.8cm}
	\caption{Steps in image segmentation of guidewire-phantom interaction and tracking contact points: aortic arch phantom (a), phantom boundaries mask (b), navigated guidewire in RCCA (c), guidewire mask (dilated for visual purpose) (d), contact and tip detection on GW (e), contact detection (f), segmented guidewire and vessel displayed on the original frame (g) and (h).
	}
	\label{ImgSegment}
\end{figure*}

Image segmentation is divided into two main phases: extracting mask images of the tool and vessel boundaries, followed by search and tracking (Fig.~\ref{search}). A Gaussian filter (Gaussian kernel of size=5) is applied to filter out noise and then, a canny-edge detector algorithm (kernel size of 3 for the Sobel operations, ratio of lower to upper threshold of 5:1) is used to extract vessel boundaries \cite{Canny, CannyGon}. To extract catheter pixels, a thresholding operation is implemented on the grayscale frame. Then, dilation followed by erosion are applied to fill possible gaps and connect lines. The imaging algorithm is tested on the cannulation of aortic arteries by a guidewire (experimental setup is explained in later sections). Fig. \ref{ImgSegment} (b) and (d) display images of the phantom boundary mask and a guidewire mask, respectively.

An algorithm with a moving search window normal to the guidewire curve is designed to sweep the entire length of the guidewire, as depicted in Fig.~\ref{search}. The algorithm moves a rectangle search window along the tangent to the guidewire ~$ \theta(s)=\frac{dy}{dx} $, where $ s $ is the position along the curve. The starting point of the search is determined with an initial search step over the base frame. The search window finds guidewire pixels and computes their centroid $C_{G}$. In each window, a parallel logic operation is made on the masked vessel image to find the point of contact with boundaries (VB) where the distance of the guidewire centerline and VB, $ C_V $, is less than a predefined contact distance. The search window position is updated along~$ \theta(s)$ once each centroid point is found and moved toward the tip. The tip is detected where the number of contiguous pixels is less than a predefined amount. Algorithm~\ref{imaging} presents the pseudo-code of the extraction and moving window search method in segmentation and tracking. The proposed moving window search is robust to possible gaps or missing points.
\begin{algorithm}[bt]
	\small
	\caption{Segmentation and tracking}
	\label{imaging}
	\textbf{Function} image processing
	\begin{algorithmic}[2]
		\State video.read($frame$)
		\State $frame \gets RGBtoGry(frame)$ 
		\State $frame \gets Gaussian filter(frame)$      \Comment{filter out noise}
		\State $B_{VS} \gets canny(frame)$            \algorithmiccomment{extract vessel boundaries}
		\While{video.read($frame$)}                        \Comment{tracking}
		\State $GW \gets threshold(frame)$       \Comment{extract guidewire}
		\State $GW \gets$ dilate and then erode GW 
		\While{Moving search window is true}                   \Comment{search}
		\For{search pixel of window normal to catheter}
		\If { pixel value is 0 in GW} 
		\State compute center of pixels 
		\EndIf
		\State \textbf{end if}
		\EndFor
		\State \textbf{end for}
		\State $C_{GW} \gets [centroid(x), centroid(y)]$
		\If {$C_{GW}-B_{VS} < predef$ } \Comment{contact detection}
		\State $P_{CP_{i}} \gets  [centroid(x), centroid(y), L_{CP_i}]$
		\EndIf
		\State \textbf{end if}
		\If {$npixel\leq minNPixel$}    \hfill  \Comment{tip detection}
		\State $P_t \gets [centroid(x), centroid(y), L_{tip}]$
		\Else
		\State $\theta \gets$ update the tangent to the catheter
		\State $P_{MovingWindow} \gets$ update and move search window
		\EndIf
		\State \textbf{end if}
		\EndWhile 
		\State \textbf{end while}
		\EndWhile
	\end{algorithmic}
	\Return {$C_{GW}, B_{VS}, P_{CP_{i}}, P_t$} 
\end{algorithm}

Fig.~\ref{ImgSegment} shows step by step segmentation results for interaction of a guidewire with the aortic arch phantom in our experimental. Fig.~\ref{ImgSegment} (g) and (f) depict segmented guidewire-vessel interaction overlaid on original frames, which show the effectiveness of the segmentation and search algorithm. 

The last step converts pixel values to length units using camera calibration parameters. Afterward, the length of any desired point along the guidewire can be computed by numerical integration. Derived parameters are: tip pose $ P_{t}=\{x_t, y_t, l_t\} $, contact point pose~$\{P_{CPi}\}=\{x_{CP_i}, y_{CP_i}, l_{CP_i}\}$ and GW center-line pose~$\{C_{GW}\}$.

\subsection{Finite element model}
\begin{figure}[b!]
	\centering
	\tikzset{every picture/.style={line width=0.75pt}} 
	\begin{tikzpicture}[x=0.75pt,y=0.75pt,yscale=-0.77,xscale=0.77]
	
	\draw [color={rgb, 255:red, 0; green, 0; blue, 0 }  ,draw opacity=1 ][line width=0.75]  (170,318.24) -- (290,318.24)(182,201.24) -- (182,331.24) (283,313.24) -- (290,318.24) -- (283,323.24) (177,208.24) -- (182,201.24) -- (187,208.24)  ;
	\draw  [color={rgb, 255:red, 128; green, 128; blue, 128 }  ,draw opacity=1 ][fill={rgb, 255:red, 155; green, 155; blue, 155 }  ,fill opacity=0.19 ] (232.28,253.34) -- (402.03,268.19) -- (398.53,308.19) -- (228.78,293.34) -- cycle ;
	\draw [color={rgb, 255:red, 128; green, 128; blue, 128 }  ,draw opacity=1 ] [dash pattern={on 4.5pt off 4.5pt}]  (230.4,272.56) -- (400.4,288.96) ;
	\draw    (266.5,206.49) .. controls (332.37,204.28) and (389.67,195.9) .. (446.5,162.99) ;
	\draw    (261.14,169.73) .. controls (327.01,167.52) and (363,162.57) .. (422.48,134) ;
	\draw    (266.5,206.49) -- (261.14,169.73) ;
	\draw    (446.5,162.99) -- (422.48,134) ;
	\draw [color={rgb, 255:red, 128; green, 128; blue, 128 }  ,draw opacity=1 ] [dash pattern={on 4.5pt off 4.5pt}]  (264.37,187.93) .. controls (330.24,185.72) and (377.93,178.03) .. (436.24,147.79) ;
	\draw [color={rgb, 255:red, 139; green, 87; blue, 42 }  ,draw opacity=1 ][line width=0.75]    (231.86,270.38) -- (260.89,190.93) ;
	\draw [shift={(261.92,188.11)}, rotate = 470.07] [fill={rgb, 255:red, 139; green, 87; blue, 42 }  ,fill opacity=1 ][line width=0.08]  [draw opacity=0] (8.93,-4.29) -- (0,0) -- (8.93,4.29) -- cycle    ;
	\draw [color={rgb, 255:red, 139; green, 87; blue, 42 }  ,draw opacity=1 ][line width=0.75]    (400.22,287.03) -- (431.91,151.42) ;
	\draw [shift={(432.59,148.5)}, rotate = 463.15] [fill={rgb, 255:red, 139; green, 87; blue, 42 }  ,fill opacity=1 ][line width=0.08]  [draw opacity=0] (8.93,-4.29) -- (0,0) -- (8.93,4.29) -- cycle    ;
	\draw [color={rgb, 255:red, 74; green, 74; blue, 74 }  ,draw opacity=1 ] [dash pattern={on 0.84pt off 2.51pt}]  (403.4,248.84) -- (397.03,329.02) ;
	\draw [color={rgb, 255:red, 74; green, 74; blue, 74 }  ,draw opacity=1 ] [dash pattern={on 0.84pt off 2.51pt}]  (233.97,232.78) -- (226.84,312.34) ;
	\draw [color={rgb, 255:red, 74; green, 74; blue, 74 }  ,draw opacity=1 ] [dash pattern={on 0.84pt off 2.51pt}]  (253.5,120.99) -- (266.5,206.49) ;
	\draw [color={rgb, 255:red, 74; green, 74; blue, 74 }  ,draw opacity=1 ] [dash pattern={on 0.84pt off 2.51pt}]  (270.5,119.49) -- (263.82,188.11) ;
	\draw [color={rgb, 255:red, 74; green, 74; blue, 74 }  ,draw opacity=1 ] [dash pattern={on 0.84pt off 2.51pt}]  (439.5,88.82) -- (434.49,147.83) ;
	\draw [color={rgb, 255:red, 74; green, 74; blue, 74 }  ,draw opacity=1 ] [dash pattern={on 0.84pt off 2.51pt}]  (397.5,99.32) -- (434.49,147.83) ;
	\draw [color={rgb, 255:red, 139; green, 87; blue, 42 }  ,draw opacity=1 ]   (256.38,120) .. controls (260.01,118.69) and (265.23,116.97) .. (269.79,118.7) ;
	\draw [shift={(253.5,120.99)}, rotate = 344.74] [fill={rgb, 255:red, 139; green, 87; blue, 42 }  ,fill opacity=1 ][line width=0.08]  [draw opacity=0] (8.93,-4.29) -- (0,0) -- (8.93,4.29) -- cycle    ;
	\draw  [color={rgb, 255:red, 208; green, 2; blue, 27 }  ,draw opacity=1 ][fill={rgb, 255:red, 225; green, 19; blue, 19 }  ,fill opacity=1 ] (432.59,148.5) .. controls (432.59,147.45) and (433.44,146.6) .. (434.49,146.6) .. controls (435.54,146.6) and (436.39,147.45) .. (436.39,148.5) .. controls (436.39,149.55) and (435.54,150.4) .. (434.49,150.4) .. controls (433.44,150.4) and (432.59,149.55) .. (432.59,148.5) -- cycle ;
	\draw  [color={rgb, 255:red, 208; green, 2; blue, 27 }  ,draw opacity=1 ][fill={rgb, 255:red, 225; green, 19; blue, 19 }  ,fill opacity=1 ] (261.92,188.11) .. controls (261.92,187.06) and (262.77,186.21) .. (263.82,186.21) .. controls (264.87,186.21) and (265.72,187.06) .. (265.72,188.11) .. controls (265.72,189.16) and (264.87,190.01) .. (263.82,190.01) .. controls (262.77,190.01) and (261.92,189.16) .. (261.92,188.11) -- cycle ;
	\draw  [color={rgb, 255:red, 128; green, 128; blue, 128 }  ,draw opacity=1 ][fill={rgb, 255:red, 128; green, 128; blue, 128 }  ,fill opacity=1 ] (398.32,288.93) .. controls (398.32,287.88) and (399.17,287.03) .. (400.22,287.03) .. controls (401.27,287.03) and (402.12,287.88) .. (402.12,288.93) .. controls (402.12,289.98) and (401.27,290.83) .. (400.22,290.83) .. controls (399.17,290.83) and (398.32,289.98) .. (398.32,288.93) -- cycle ;
	\draw  [color={rgb, 255:red, 128; green, 128; blue, 128 }  ,draw opacity=1 ][fill={rgb, 255:red, 128; green, 128; blue, 128 }  ,fill opacity=1 ] (228.5,272.56) .. controls (228.5,271.51) and (229.35,270.66) .. (230.4,270.66) .. controls (231.45,270.66) and (232.3,271.51) .. (232.3,272.56) .. controls (232.3,273.61) and (231.45,274.46) .. (230.4,274.46) .. controls (229.35,274.46) and (228.5,273.61) .. (228.5,272.56) -- cycle ;
	\draw [color={rgb, 255:red, 139; green, 87; blue, 42 }  ,draw opacity=1 ]   (404.6,96.17) .. controls (417.62,87.04) and (422.67,87.19) .. (433.24,89.56) ;
	\draw [shift={(402,98.04)}, rotate = 323.75] [fill={rgb, 255:red, 139; green, 87; blue, 42 }  ,fill opacity=1 ][line width=0.08]  [draw opacity=0] (8.93,-4.29) -- (0,0) -- (8.93,4.29) -- cycle    ;
	\draw [color={rgb, 255:red, 74; green, 74; blue, 74 }  ,draw opacity=1 ][line width=0.75]    (436.24,147.29) -- (471,147.93) ;
	\draw [shift={(474,147.99)}, rotate = 181.06] [fill={rgb, 255:red, 74; green, 74; blue, 74 }  ,fill opacity=1 ][line width=0.08]  [draw opacity=0] (8.93,-4.29) -- (0,0) -- (8.93,4.29) -- cycle    ;
	\draw [color={rgb, 255:red, 74; green, 74; blue, 74 }  ,draw opacity=1 ][line width=0.75]    (434.49,146.1) -- (434.04,114.99) ;
	\draw [shift={(434,111.99)}, rotate = 449.17] [fill={rgb, 255:red, 74; green, 74; blue, 74 }  ,fill opacity=1 ][line width=0.08]  [draw opacity=0] (8.93,-4.29) -- (0,0) -- (8.93,4.29) -- cycle    ;
	\draw [color={rgb, 255:red, 74; green, 74; blue, 74 }  ,draw opacity=1 ][line width=0.75]    (224.16,187.41) -- (258.92,188.05) ;
	\draw [shift={(261.92,188.11)}, rotate = 181.06] [fill={rgb, 255:red, 74; green, 74; blue, 74 }  ,fill opacity=1 ][line width=0.08]  [draw opacity=0] (8.93,-4.29) -- (0,0) -- (8.93,4.29) -- cycle    ;
	\draw [color={rgb, 255:red, 74; green, 74; blue, 74 }  ,draw opacity=1 ][line width=0.75]    (263.82,186.21) -- (264.64,150.81) ;
	\draw [shift={(264.71,147.81)}, rotate = 451.33] [fill={rgb, 255:red, 74; green, 74; blue, 74 }  ,fill opacity=1 ][line width=0.08]  [draw opacity=0] (8.93,-4.29) -- (0,0) -- (8.93,4.29) -- cycle    ;
	\draw    (498.3,268.47) .. controls (505.31,265.66) and (508.56,263.94) .. (517.13,259.18) ;
	\draw    (548.41,296.97) -- (517.13,259.18) ;
	\draw [color={rgb, 255:red, 128; green, 128; blue, 128 }  ,draw opacity=1 ] [dash pattern={on 4.5pt off 4.5pt}]  (497.85,299.09) .. controls (513.42,291.2) and (521.85,285.54) .. (532.85,277.55) ;
	\draw [color={rgb, 255:red, 74; green, 74; blue, 74 }  ,draw opacity=1 ] [dash pattern={on 0.84pt off 2.51pt}]  (538.19,217.14) -- (529,320) ;
	\draw [color={rgb, 255:red, 7; green, 54; blue, 110 }  ,draw opacity=1 ]   (492,229) -- (559.09,309.98) ;
	\draw    (519.92,315.22) .. controls (534.58,307.54) and (537.22,305.11) .. (548.41,296.97) ;
	\draw  [color={rgb, 255:red, 128; green, 128; blue, 128 }  ,draw opacity=1 ] (530.57,272.89) -- (537.15,269.28) -- (539.44,273.93) -- (532.85,277.55) -- cycle ;
	\draw [color={rgb, 255:red, 128; green, 128; blue, 128 }  ,draw opacity=1 ][fill={rgb, 255:red, 155; green, 155; blue, 155 }  ,fill opacity=1 ] [dash pattern={on 4.5pt off 4.5pt}]  (532.85,277.55) -- (569.3,258.31) ;
	\draw [color={rgb, 255:red, 65; green, 117; blue, 5 }  ,draw opacity=1 ][line width=0.75]    (505.72,219.02) -- (549,313) ;
	\draw [color={rgb, 255:red, 74; green, 74; blue, 74 }  ,draw opacity=1 ]   (522.63,248.51) .. controls (528.43,245.26) and (529.95,244.71) .. (536.21,245.03) ;
	\draw [shift={(520,250)}, rotate = 330.58] [fill={rgb, 255:red, 74; green, 74; blue, 74 }  ,fill opacity=1 ][line width=0.08]  [draw opacity=0] (8.93,-4.29) -- (0,0) -- (8.93,4.29) -- cycle    ;
	\draw [color={rgb, 255:red, 74; green, 74; blue, 74 }  ,draw opacity=1 ]   (504.53,238.14) .. controls (518.45,228.05) and (523.23,227.35) .. (538.04,228.97) ;
	\draw [shift={(502,240)}, rotate = 323.33] [fill={rgb, 255:red, 74; green, 74; blue, 74 }  ,fill opacity=1 ][line width=0.08]  [draw opacity=0] (8.93,-4.29) -- (0,0) -- (8.93,4.29) -- cycle    ;
	\draw    (507.03,294.07) .. controls (506.22,278.71) and (500.12,284.09) .. (498.3,268.47) ;
	\draw    (519.92,315.22) .. controls (519.24,300.14) and (509.46,309.28) .. (507.03,294.07) ;
	\draw [color={rgb, 255:red, 155; green, 155; blue, 155 }  ,draw opacity=1 ]   (498,224) -- (490.92,238.57) ;
	\draw [color={rgb, 255:red, 74; green, 74; blue, 74 }  ,draw opacity=1 ]   (491.12,223.95) .. controls (495.87,219.54) and (498.55,218.07) .. (504.22,217.21) ;
	\draw [shift={(488.94,226.03)}, rotate = 315.83] [fill={rgb, 255:red, 74; green, 74; blue, 74 }  ,fill opacity=1 ][line width=0.08]  [draw opacity=0] (8.93,-4.29) -- (0,0) -- (8.93,4.29) -- cycle    ;
	\draw [color={rgb, 255:red, 155; green, 155; blue, 155 }  ,draw opacity=1 ]   (547,231) -- (535.21,242.03) ;
	\draw   (460.13,188.38) -- (453.41,185.49) -- (454.23,184.6) -- (446.63,177.58) -- (448.28,175.78) -- (455.89,182.8) -- (456.72,181.91) -- cycle ;
	\draw [color={rgb, 255:red, 74; green, 74; blue, 74 }  ,draw opacity=1 ]   (441.6,137.53) .. controls (451.77,143.87) and (455.63,150.93) .. (450,163) ;
	\draw [shift={(439,136)}, rotate = 29.2] [fill={rgb, 255:red, 74; green, 74; blue, 74 }  ,fill opacity=1 ][line width=0.08]  [draw opacity=0] (8.93,-4.29) -- (0,0) -- (8.93,4.29) -- cycle    ;
	\draw [color={rgb, 255:red, 74; green, 74; blue, 74 }  ,draw opacity=1 ]   (248.33,197.5) .. controls (241.93,187.38) and (241.31,178.99) .. (249,172) ;
	\draw [shift={(250,200)}, rotate = 234.86] [fill={rgb, 255:red, 74; green, 74; blue, 74 }  ,fill opacity=1 ][line width=0.08]  [draw opacity=0] (8.93,-4.29) -- (0,0) -- (8.93,4.29) -- cycle    ;
	\draw (220.88,210.33) node  [font=\footnotesize,color={rgb, 255:red, 139; green, 87; blue, 42 }  ,opacity=1 ]  {$(u_{xi\ },u_{yi})$};
	\draw (352.5,222.74) node  [font=\footnotesize,color={rgb, 255:red, 139; green, 87; blue, 42 }  ,opacity=1 ]  {$(u_{x( i+1) \ },u_{y(i+1)})$};
	\draw (215.8,272.8) node  [font=\small]  {$N_{i}$};
	\draw (419.7,289.04) node  [font=\small]  {$N_{i+1}$};
	\draw (409.5,78.5) node  [font=\footnotesize,color={rgb, 255:red, 139; green, 87; blue, 42 }  ,opacity=1 ]  {$\varphi _{i+1}$};
	\draw (259.5,102.9) node  [font=\footnotesize,color={rgb, 255:red, 139; green, 87; blue, 42 }  ,opacity=1 ]  {$\varphi _{i}$};
	\draw (465.5,114.74) node  [font=\footnotesize]  {$F_{y( i+1)}$};
	\draw (500,152.74) node  [font=\footnotesize]  {$F_{x( i+1)}$};
	\draw (243.07,151.24) node  [font=\footnotesize]  {$F_{yi}$};
	\draw (224.5,176.74) node  [font=\footnotesize]  {$F_{xi}$};
	\draw (183,188.24) node  [font=\small]  {$y$};
	\draw (301,316.24) node  [font=\small]  {$x$};
	\draw (519,220) node  [font=\footnotesize,color={rgb, 255:red, 0; green, 0; blue, 0 }  ,opacity=1 ]  {$\varphi $};
	\draw (558.91,226.35) node  [font=\footnotesize,color={rgb, 255:red, 0; green, 0; blue, 0 }  ,opacity=1 ]  {$-\frac{dw}{dx}$};
	\draw (485,245.5) node  [font=\footnotesize,color={rgb, 255:red, 0; green, 0; blue, 0 }  ,opacity=1 ]  {$\gamma _{xy}$};
	\draw (497,310) node  [font=\tiny] [align=left] {Neutral \\axis};
	\draw (569.5,288.67) node  [font=\tiny,color={rgb, 255:red, 3; green, 59; blue, 125 }  ,opacity=1 ] [align=left] {Timoshenko};
	\draw (560.5,317) node  [font=\tiny,color={rgb, 255:red, 65; green, 117; blue, 5 }  ,opacity=1 ] [align=left] {Euler-Bernoulli};

	\end{tikzpicture} 
	\caption{Generic beam element, including the nodal forces ($ F $) and the displacement vectors ($ u $), showing initial and deformed states. Euler-Bernoulli beam is compared with that of a Timoshenko. }
	\label{element}
\end{figure}
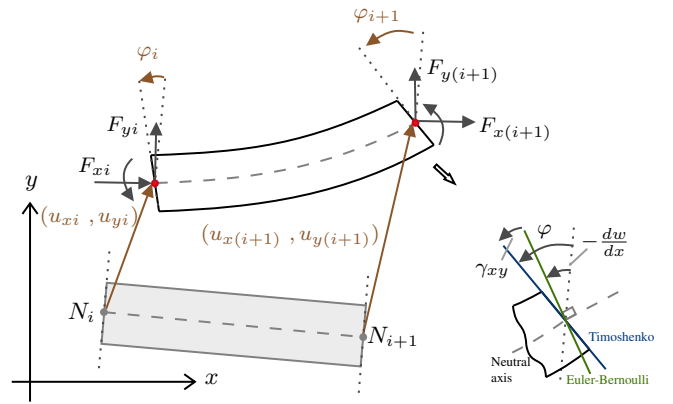

Consider a planar two-node beam element of length $ l $, where each node has three degrees of freedom, as defined in Fig. \ref{element}. The nodal displacement vector contains longitudinal ($ u_x $), transverse ($ u_y $) displacement and rotation ($ \varphi $) for both nodes. 
\begin{equation}
\label{NodalDOF}
e = \begin{bmatrix}
u_{x1}&u_{y1}&\varphi_{1}&u_{x2}&u_{y2}&\varphi_{2}
\end{bmatrix}^{T} 
\end{equation}
The global elements coordinate vector is defined as
\begin{equation}
\label{elementDOF}
\footnotesize
\left \{e\right \}\equiv
\begin{bmatrix}
u_{x1}&u_{y1}&\varphi_{1}&...&u_{x(N+1)}&u_{y(N+1)}&\varphi_{(N+1)}
\end{bmatrix}^{T} \in R^{n}
\end{equation}
where $ N $ is the number of element and $ n=3(N+1) $ is the number of degrees of freedom.

In the earlier study \cite{razban2018sensor}, we presented results based on Euler-Bernoulli (EB) beam theory whereas here the formulation has been updated to the Timoshenko beam theory. EB assumes thin beams where angular distortion due to shear deformation is considered negligible, and cross-section is perpendicular to the bending line (neutral axis). Timoshenko beam theory is generic compared to EB theory. Its employed formulation accounts for large axial, bending, and shear deformations as well as large translation and rotation of the beam structure. 
The governing equations of Timoshenko beam theory are the following:
\begin{equation}
\label{Timoshenko}
\frac{d^{2}}{dx^{2}}(EI\frac{d\varphi}{dx})=q(x)
\end{equation}
\begin{equation}
\label{Timoshenko2}
\frac{dw}{dx}=\varphi -\frac{1}{\kappa AG}\frac{d}{dx}(EI\frac{d\varphi}{dx})
\end{equation}
where $ E $ and $ G $ are the elastic modulus and shear modulus. $ L $, $ I $ and $ A $ are the length of the beam, second moment of area and cross section area, respectively. $ \kappa $ is the Timoshenko shear coefficient and $ q(x) $ is the load \cite{timoshenko1922x}.
Fig. \ref{element} compares Timoshenko beam deformation to EB, where the cross-section of a beam element remains planed but not necessarily perpendicular to the beam axis, i.e., $ \varphi(x)\neq \frac{dw}{dx} $ and $\gamma _{xy}$ is shear \cite{thomas1973timoshenko, reddy1999dynamic}. Timoshenko can be used for thick as well as slender beams, which is the more accurate choice for modeling of catheters and guidewires, especially where the beam cross-sectional dimensions is small compared to typical distances along its axis.

The equilibrium of a finite element beam model at time $ t $ is
\begin{equation}
\label{equilibrium}
M ~^t\!\ddot{u}+C ~^t\!\dot{u}=~^t\!R - ~^t\!F
\end{equation}
where $ M \in R^{n \times n} $ is the mass matrix and $ C=\alpha M $ is the damping matrix proportional to mass with coefficient $\alpha$. $ R \in R^{n} $ is the external load vector and $ F \in R^{n}$ is the internal force vector in global coordinates. $ \ddot{u} $ and $ \dot{u} $ are acceleration and velocity vectors, respectively \cite{bathe1975finite}. The velocities and acceleration vectors are small during navigation because of continuous contact with the vessel. Further, considering the low mass of the catheter/guidewire, the dynamic force is small compared to large external forces caused by large deflections. Thus, we are implementing a quasi-static solution to minimize the computational cost and reach a real-time execution.

Nodal force is formed based on individual vectors considering element connectivity. 
\begin{equation}
\label{KU}
\prescript{t}{}F_j=K ~^t\!u_j
\end{equation}
\begin{equation}
\label{K}
K\equiv K_L+K_{NL}
\end{equation}
$\prescript{t}{}F_j \in R^6 $ is the internal nodal point force of the element $ j $ and $\prescript{t}{}u_j \in R^{6} $ is the displacement vector. The stiffness matrix ($ K \in R^{6\times6} $) consists of linear ($ K_L \in R^{6\times6} $) and nonlinear ($ K_{NL} \in R^{6\times6}$) parts \cite{ bathe1982finite}. Strain-displacement equations contain nonlinear terms that must be considered in the nonlinear part of the stiffness matrix. $ K_{NL} $ is achieved by applying Castigliano's theorem to the strain energy which counts the interaction between axial load and lateral deformation \cite{przemieniecki1968theory}. 

After transferring the finite element matrix of the local principal axis of the elements to global cartesian co-ordinate and performing an element assemblage process, the incremental equilibrium equation of quasi-static analysis is:
 \begin{equation}
 \label{}
(\prescript{t}{t}{K}_L + \prescript{t}{t}{K}_{NL}) \prescript{\Delta t}{}u=\prescript{t+\Delta t}{}{R} - \prescript{t}{t}{F}
\end{equation}
where $\prescript{t}{t}{K}_L$ and $\prescript{t}{t}{K}_{NL}$ are stiffness matrices at time $ t $, $ \prescript{t}{t}{F }$ is the nodal point force at time $ t $ and $\prescript{t+\Delta t}{}{R} $ is the externally applied nodal load at time $ t+\Delta t $ \cite{bathe1979large}. 
The case of our problem is an inverse finite element since we have nodal displacement vectors $ u_i $ from imaging data to calculate $ R_i $ as the interaction contact forces. $ R $ is the resultant of all external forces applied to the catheter from the vessel, including normal and friction forces. 
\begin{equation}
\label{R}
~^t\!R=~^t\!R_f + ~^t\!R_n
\end{equation}
\begin{algorithm}[b]
	\caption{Model preparation}
	\label{Model}
	\small 
	\begin{algorithmic}[2]
		\State   \textbf{Input} Data from image-based pose measurement 
		\State 	$ L_{beam} \gets L_{tip} $
		\State Create cantilever beam from intrinsic shape and $ L_{beam} $ 
		\For {$ i < N_{CP} $}
		\State  $ L_{BC_i} \gets L_{CP_i} $
		\State Locate position of $ P_{BC_i} $ on cantilever beam based on $ L_{BC_i} $
		\State $ d_i=P_{CP_i}- P_{BC_i}$
		\State Apply $ d_i $ to $ BC_i $
		\EndFor
		\For {$ n<N_{elements} $}
		\State Compute element length from tip $ S_{e} $
		\State $ EI_{element} \gets EI(S_e) $ \Comment assign mechanical property 
		\EndFor
	\end{algorithmic}
\end{algorithm}
$ R $ is computed regardless of type, condition, or material properties of the external environment. Viscoelastic effects of the tissue are already seen in the nodal displacement $ u $. Consequently, no information about contact condition and tissue characteristics is needed. A geometrical computation is essential to create the cantilever model from imaging data, as given in Algorithm \ref{Model}. The model is made based on intrinsic shape information and total length of guidewire used in experiments, $ L_{tip} $. The boundary conditions are located on the model at the points (with $ L_{BC} $) having an equal length to the contact points $ L_{CP} $. The displacement vector $ d_i $ is measured from $ BC_i $ to $ CP_i $ (see \ref{eq.1}) which is assigned to its associated boundary condition. Each element has an individual mechanical property $ EI $, which is elaborated in the next section. The FEM modeling is coded in C++ and the associated beam model is solved using ABAQUS, B21 beam with active $ Nlgeom $.
\subsection{Flexural rigidity distribution}
Cardiovascular devices have different mechanical properties, designs, and geometries to achieve specific tasks. The local flexural rigidity of guidewires varies along the its length. Clinical performance, i.e., steerability, torquability, penetration, deliverability, and safety, depends on the mechanical property profile over the length \cite{walker2013guidewire}. It needs lower flexural rigidity at the distal tip while higher rigidity is desired toward the proximal end.  
\begin{figure}[t]
	\centering
	\includegraphics[width=0.42\textwidth]{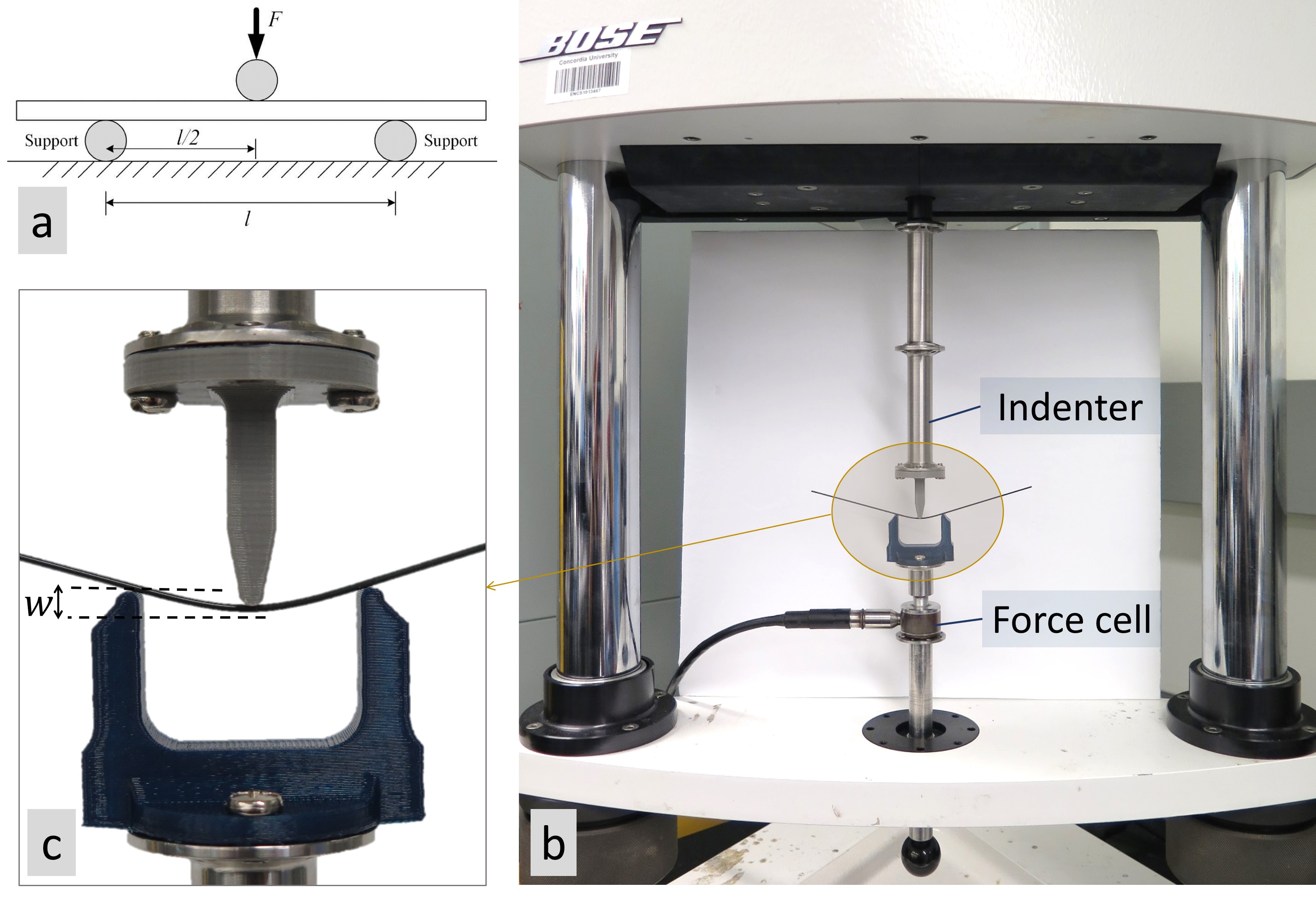}	
	\caption{Three-point bending test: schematic concept (a), Electroforce dynamic testing machine (b), deflected guidewire under test (c).}
	\label{bendingtest}
\end{figure}
\begin{figure}[t]
	\centering
	\includegraphics[width=0.38\textwidth]{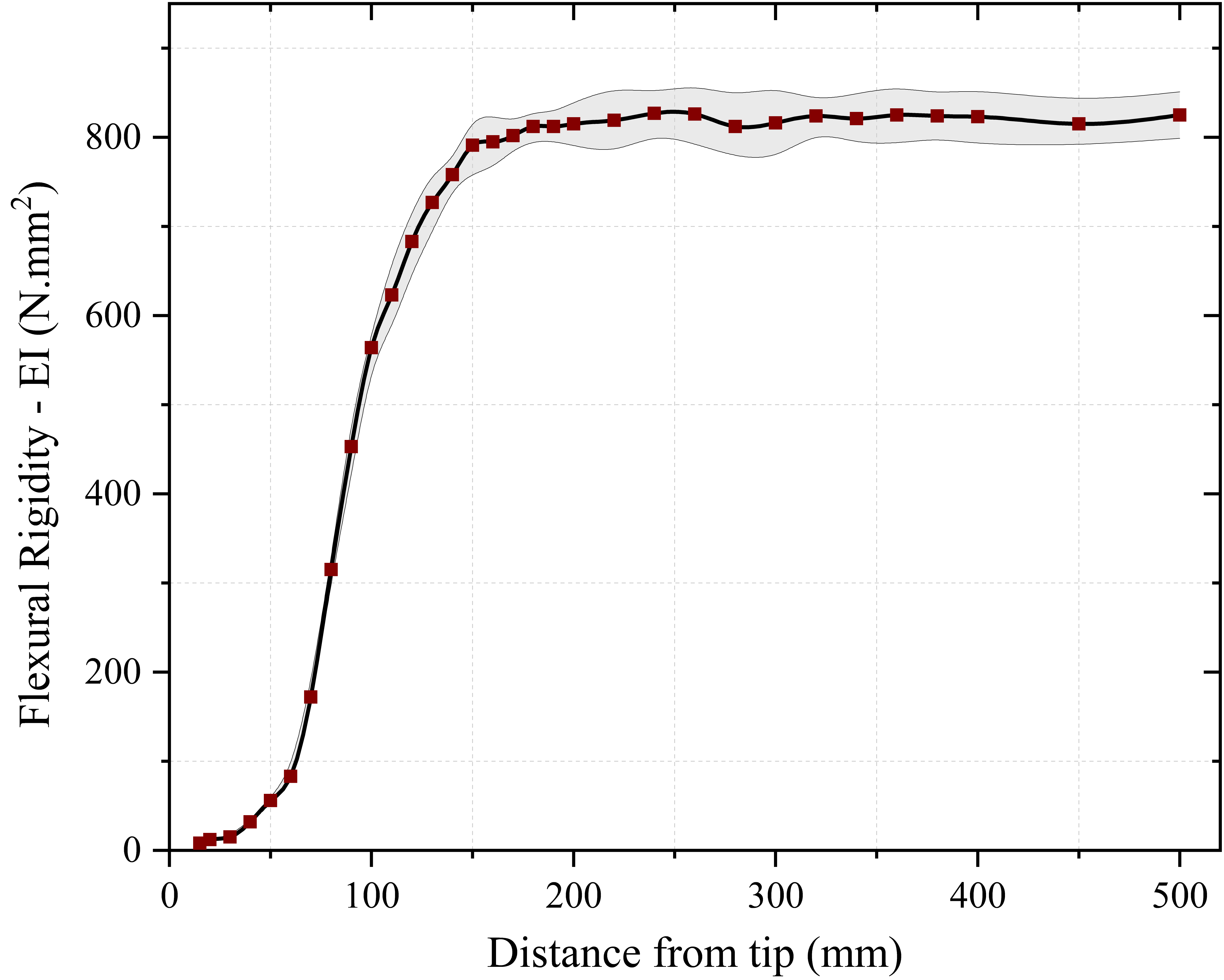}
	\caption{Flexural rigidity distribution of the guidewire based on distance from distal tip. Average value and deviation are shown.}
	\label{EIdistribution}
\end{figure}

 \begin{figure*}[]
	\centering
	\includegraphics[width=0.9\textwidth, trim={0 0 2 0cm}]{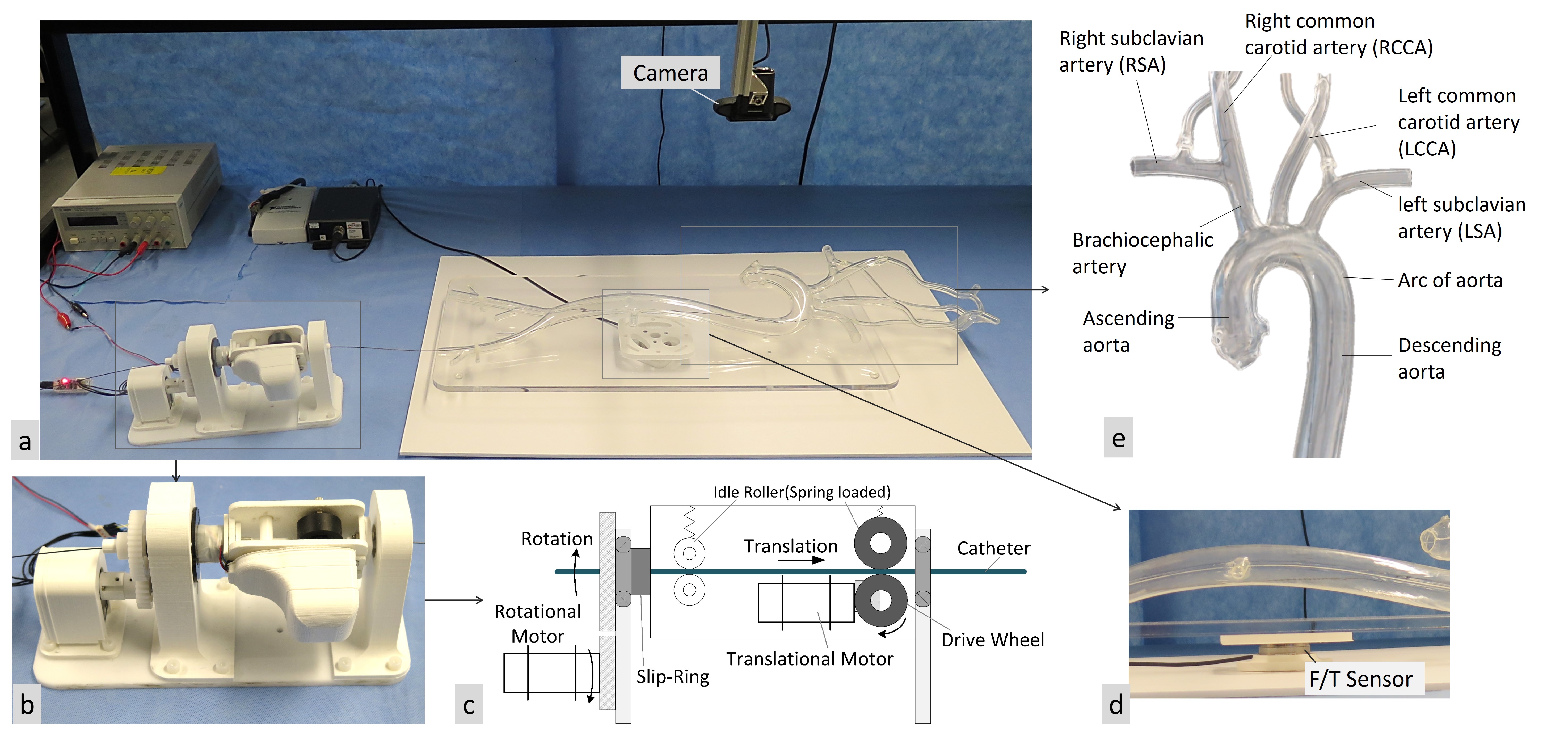}
	\captionsetup{margin=1cm}	
	\caption{Experimental setup used for force monitoring tests (a) depicting robotic catheter drive (b), schematic design (c), force measurement platform by F/T sensor (d), aortic arteries (e).}
	\label{setup}
\end{figure*}

The flexural rigidity distribution is measured through sequential three-point bending tests, Fig. \ref{bendingtest}, along the length as presented in \cite{razban2018sensor}. The bending modulus is proportional to the fourth power of the radius, so a slight change in radius can greatly affect the accuracy of bending modulus measurements, consequently force estimation. Hence, we aim to determine and utilize $ EI $ instead of $ E $. Fig.~\ref{EIdistribution} shows the flexural rigidity distribution of a Zipwire\textsuperscript{TM} Stiff guidewire (Boston Scientific, USA) measured as a function of length from the tip, $ EI(s) $. To do so, sequential three-point bending tests were performed over 0\,-\,500\,mm distal end at increments of 10\,mm (over 0\,-\,200mm), 20\,mm (over 200\,-\,400mm) and 50mm (over 400\,-\,500mm). We used a Bose® Electroforce 3200 dynamic testing machine (Bose Corp., Massachusetts, US). The support span length was 30\,mm, and loading speed of the machine was fixed at 20\,mm/min to eliminate any potential dynamic effect. Each test was repeated three times, and the average value and standard deviation were computed. Rigidity changes sharply within 15\,cm of the distal tip and stays consistent beyond it. In the model creation, the algorithm measures element lengths from the tip and assigns individual mechanical properties to the elements based on the rigidity graph (see algorithm \ref{Model} lines 9\,-\,11).
\section{Experimental platform and study protocol}
\subsection{Experimental setup}
The setup has a transparent, realistic, anthropomorphic training phantom representing the aortic arch and extended carotid structure with normal configuration (Fig. \ref{setup}). A camera is mounted above the phantom to provide image feedback for force estimation and also operator visual guidance. A robotic platform is designed to remotely navigate the guidewire (Fig. \ref{setup} (b)). The robotic driver (slave) is controlled by the operator (master) using a keyboard/joystick to command the manipulation procedure. A Zipwire\textsuperscript{TM} Stiff guidewire with angled tip shape is used with no support catheter. The operator performs navigation under image-guidance using a live gray-scale image simulating 2D fluoroscopy projected on a monitor. It is a general case study of any non-steerable catheter or guidewire navigation through cardiovascular anatomy.

Part of the aim of this paper is to study and compare the total force exerted on the vasculature, i.e., the insertion force in our setup, with intraluminal interaction contact forces. To do so, the phantom is mounted on a six-axis force-torque (F/T) sensor as shown in Fig. \ref{setup} (d) (Mini40; ATI Industrial Automation, Inc, Apex, NC). The setup provides force measurement in the X, Y and Z directions. The phantom is placed on its gravity center to hinder tilting or vibration during the procedure. The force sensor platform and phantom are isolated from any other support or contact, so that the force measurement corresponds to the resultant catheter-vessel force vector. The F/T sensor readings are obtained through a Data Acquisition system (National Instrument DAQ) using a software which records data at 60\,Hz and computes resultant magnitudes of the 3D force measurements. Measurements are zeroed at the very beginning of each test to omit the weight of the phantom or any undesired load.
 
\textit{Robotic Navigation System}: A robotic driver system is designed based on the methods in conventional manual navigation of non-steerable catheters/guidewire, i.e., techniques similar to the push, pull, and twist. The design (Fig. \ref{setup} (c)) has two degrees of freedom insertion and rotation, similar to the designs in \cite{thakur2009design}. It consists of a translational driver unit mounted on a slip-ring gantry, which allows unlimited rotational motion independent and simultaneous to the insertion motion. Simultaneous rotation-insertion motion is a acquired skill in manual manipulation that is featured in the proposed robotic system. The design includes two servomotors (Dynamixel XH430 series, ROBOTIS, CA, US) under velocity control based on a PID controller. Catheter translation motion is achieved by a fractional drive wheel and a secondary spring-loaded idler roller coupled opposite-side of the drive wheel to guarantee sufficient frictional force. The second motor rotates the whole translation unit on a housing gantry equipped with a slip-ring. It allows unlimited rotation of the catheter to facilitate maneuverability. 
    
\subsection{Study protocol and data analysis}
One operator performed remote robotic navigation of the guidewire through supra-aortic vessels, including the right subclavian artery (RSA), the right common carotid artery (RCCA), the left common carotid artery (LCCA) and the left subclavian artery (LSA), see Fig. \ref{setup} (e). Cannulations were repeated five times for each targeted artery. The position and length of contact interactions are obtained from image segmentation along with the CF estimation results from the FEM model to visualize CF monitoring on vessel boundaries. Resultant insertion forces on the phantom are recorded with the F/T sensor. The quantitative analysis is performed on the parameters of intraluminal CF and resultant forces, including: average of peak force, standard deviation\,(STD) of peak force, average of mean force, force-time impact (force integral over time). The FEM estimation error is analyzed based on GW deformed shape estimation, i.e., planar distance between actual GW and the simulated results. If ($ h_i $, $ q_i $) represent coordinates of each points on the actual shape and ($ H_i $, $ Q_i $) as the coordinates of points from the FEM model, then the estimation error ($ E_P $) is expressed by 
\begin{equation}
\label{Error}
E_{P}=\sqrt{(h_i-H_i)^{2})+(q_i-Q_i)^{2})}
\end{equation}
The structural stress of the GW is obtained from the FEM model. It results in an intraluminal stress study during navigation. 
\section{Results and Discussion}
\subsection{Intraluminal contact forces}
\begin{figure}[]
	\centering
	\begin{subfigure}{0.23\textwidth}
		\includegraphics[width=\textwidth,trim={0 0 0 0cm},clip]{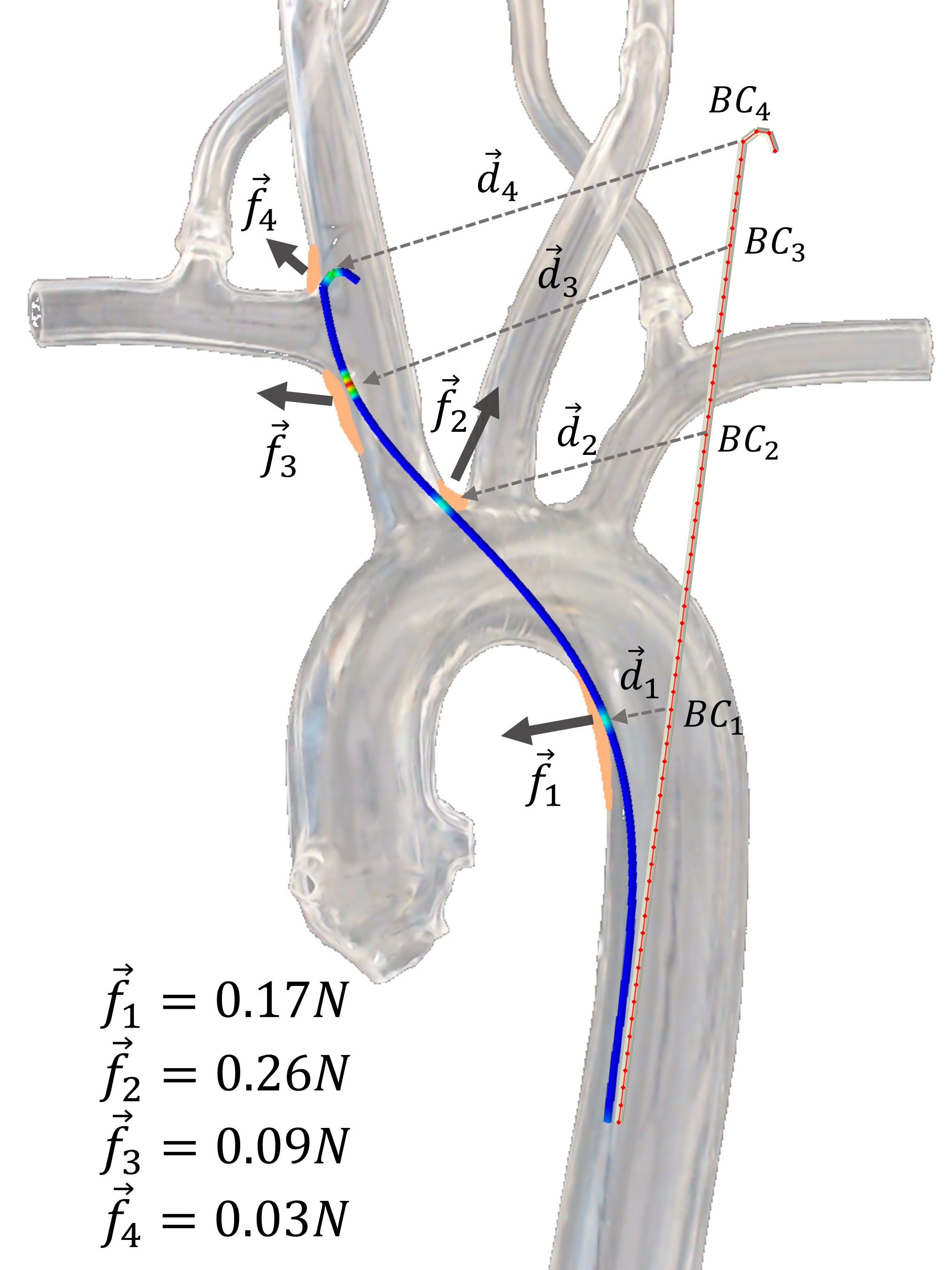}
		\caption{}
	\end{subfigure}%
	\hspace{3mm}
	\begin{subfigure}{0.23\textwidth}
		\includegraphics[width=\textwidth,trim={0 0 0 0cm},clip]{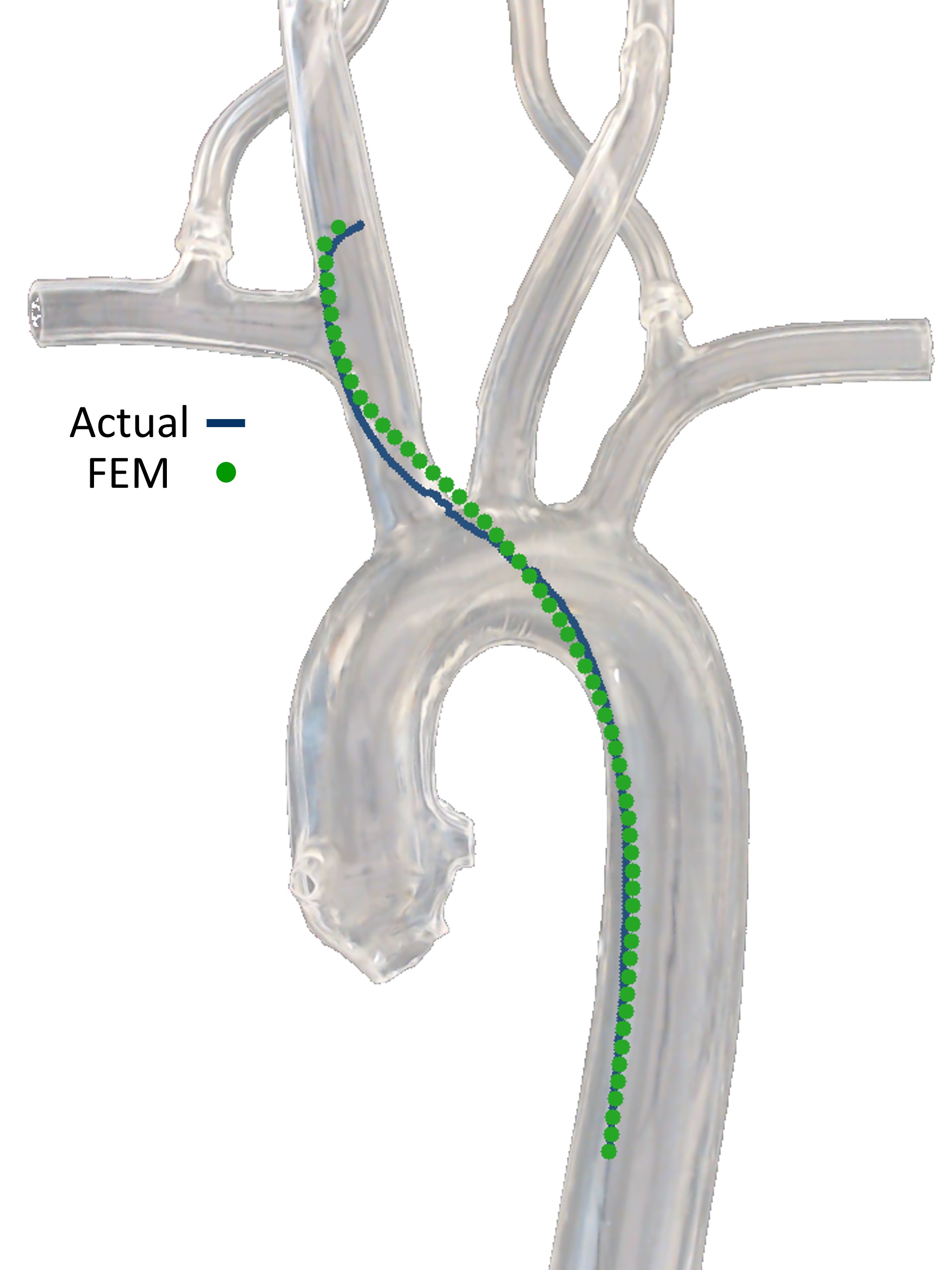}
		\caption{}  
	\end{subfigure}%
	\caption{FEM model of GW at a given time: meshed beam model, boundary conditions, deflection vectors, simulated shape and estimated CF vectors applied to the vessel wall (a), actual GW from imaging compared to FEM modeling (b).}
	\label{instantstate}
\end{figure}

Fig. \ref{instantstate} depicts an example of guidewire FEM model and simulated results at a given time during cannulation of RCCA. Subfigure (a) presents the FEM model structure: the guidewire beam model (meshed in red), deflection vector $ d_{i} $ which is applied to boundary condition $ BC_{i} $, the simulated guidewire in blue contour highlighting reaction forces, direction and magnitudes of four local contact force vectors and the contact positions on the vessel wall. In this example, contact forces are: $ \vert \vec{f_{1}} \vert=0.17\,N, \vert\vec{f_{2}}\vert=0.26\,N, \vert\vec{f_{3}}\vert=0.09\,N, \vert\vec{f_{4}}\vert=0.03\,N $. Contact forces closer to the distal end ($ \vec{f_{3}} $ and $ \vec{f_{4}} $) are significantly lower than $ \vec{f_{1}} $ and $ \vec{f_{2}} $ which is due to lower GW flexural rigidity in that region. Fig. \ref{instantstate}\,(b) compares modeling results represented as green dots to the actual GW shape from image segmentation in blue curve. A visual comparison suggests that the model mimics the GW behavior.

\begin{figure}[t] 
	\centering
	\includegraphics[width=0.42\textwidth]{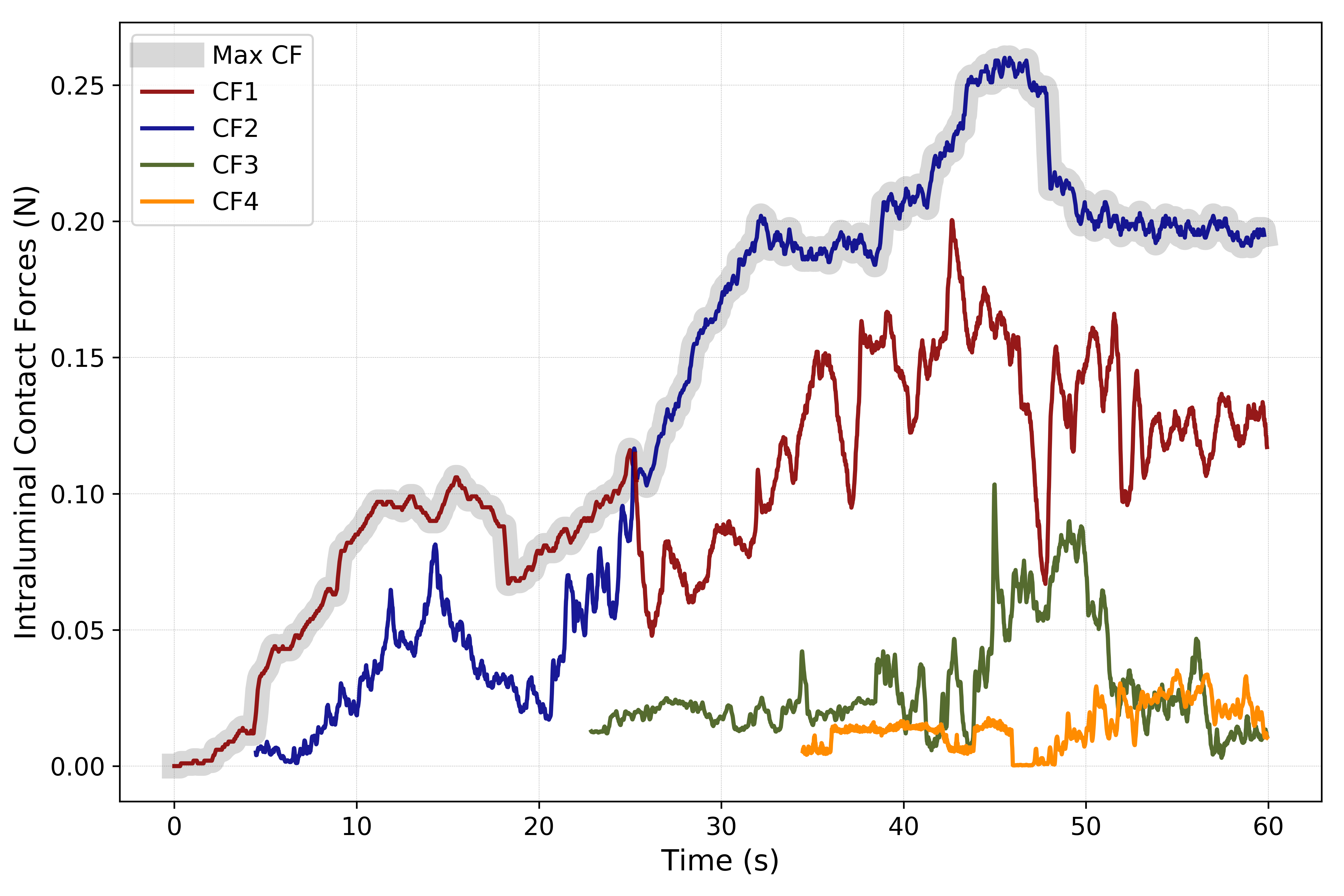}
	\caption{Example of image-based intraluminal contact forces monitoring for every local contact points through RCCA navigation. $ CF_4 $ to $ CF_1 $ are named from the distal tip toward the proximal end.}
	\label{CFsRCCA}
\end{figure}

Fig. \ref{CFsRCCA} displays a graph tracking local CFs during navigation of the RCCA. At the beginning of the procedure, the GW has only one contact, and the number of contacts is increasing to four as GW is advancing. The maximal ranges of $ CF_3 $ and $ CF_4 $ are 0.04\,N - 0.11\,N whereas the ranges of $ CF_1 $ and $ CF_2 $ are about 0.205\,N - 0.26\,N. The ascending trend is seen for all CF values, but $ CF_1 $ and $ CF_2 $ forces increase to larger values. This is because the $ CF_1 $ and $ CF_2 $ positions are moving away from the tip over insertion which gradually shifts contact to the stiffer regions, whereas $ CF_3 $ and $ CF_4 $ are just happening at the softer regions close to the tip. CF magnitudes are fluctuating because of the GW slip-stick motion on the phantom wall.
Maximal intraluminal CF, as the main safety performance metric in interventional procedures, is highlighted on a gray shadow in Fig. \ref{CFsRCCA}. Maximal CF was on $ CF_1 $ and turns to $ CF_2 $ after the guidewire gets in contact with more points and is being highly deflected at the middle region (see Fig. \ref{CFsRCCA} where $ CF_3 $ starts). It was observed that every procedure has a unique CF monitoring trend; however, maximal contact forces were always happening on contact points far from the tip rather than near.
\begin{figure}[t] 
	\centering
	\includegraphics[width=0.4\textwidth, trim={0 0.45cm 0 0.7cm}, clip]{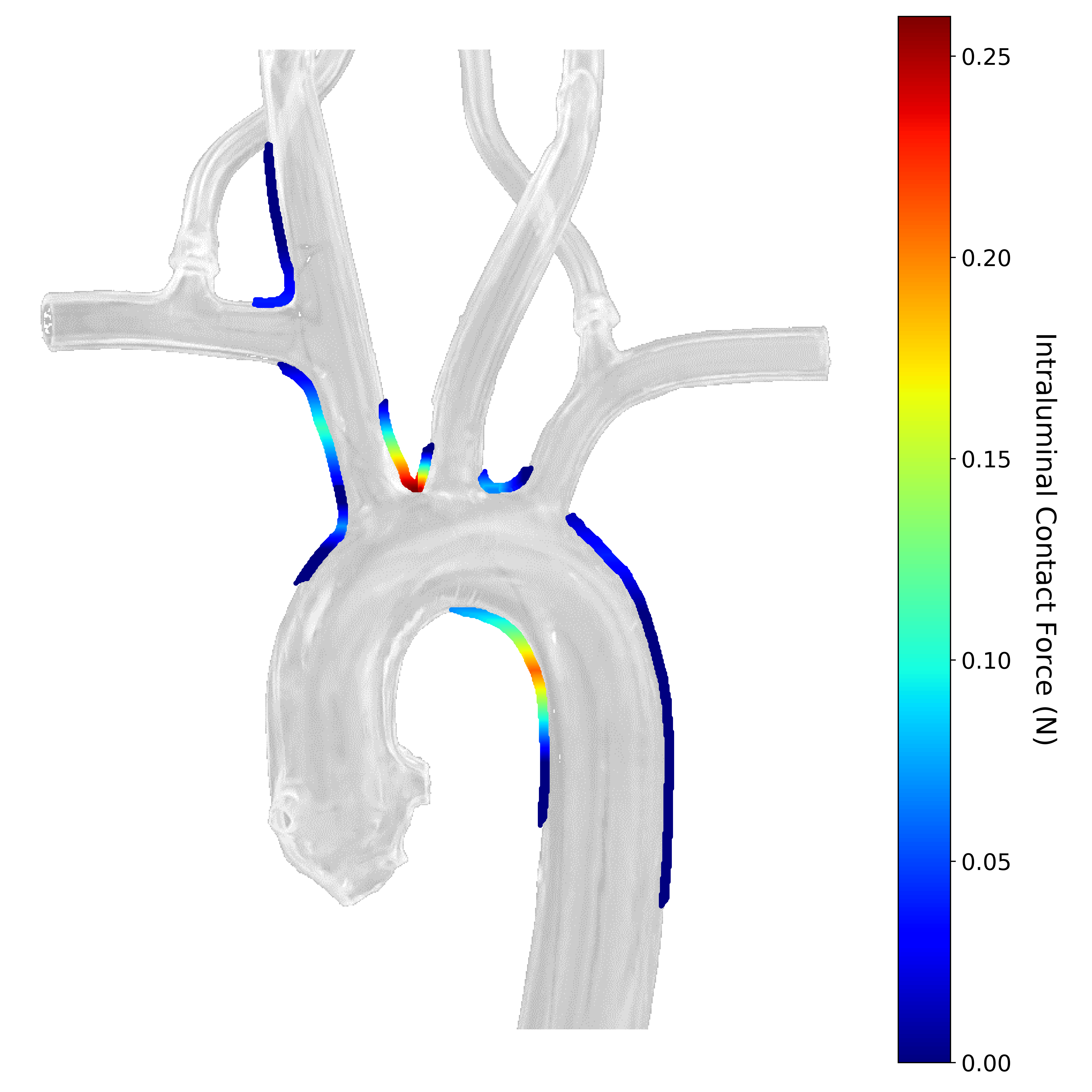}
	\caption{Guidewire-vessel interactions and intraluminal contact forces map computed during navigation of RCCA.}
	\label{RCCACountor}
\end{figure}
\begin{figure*}[] 
	\centering
	\includegraphics[width=0.32\textwidth]{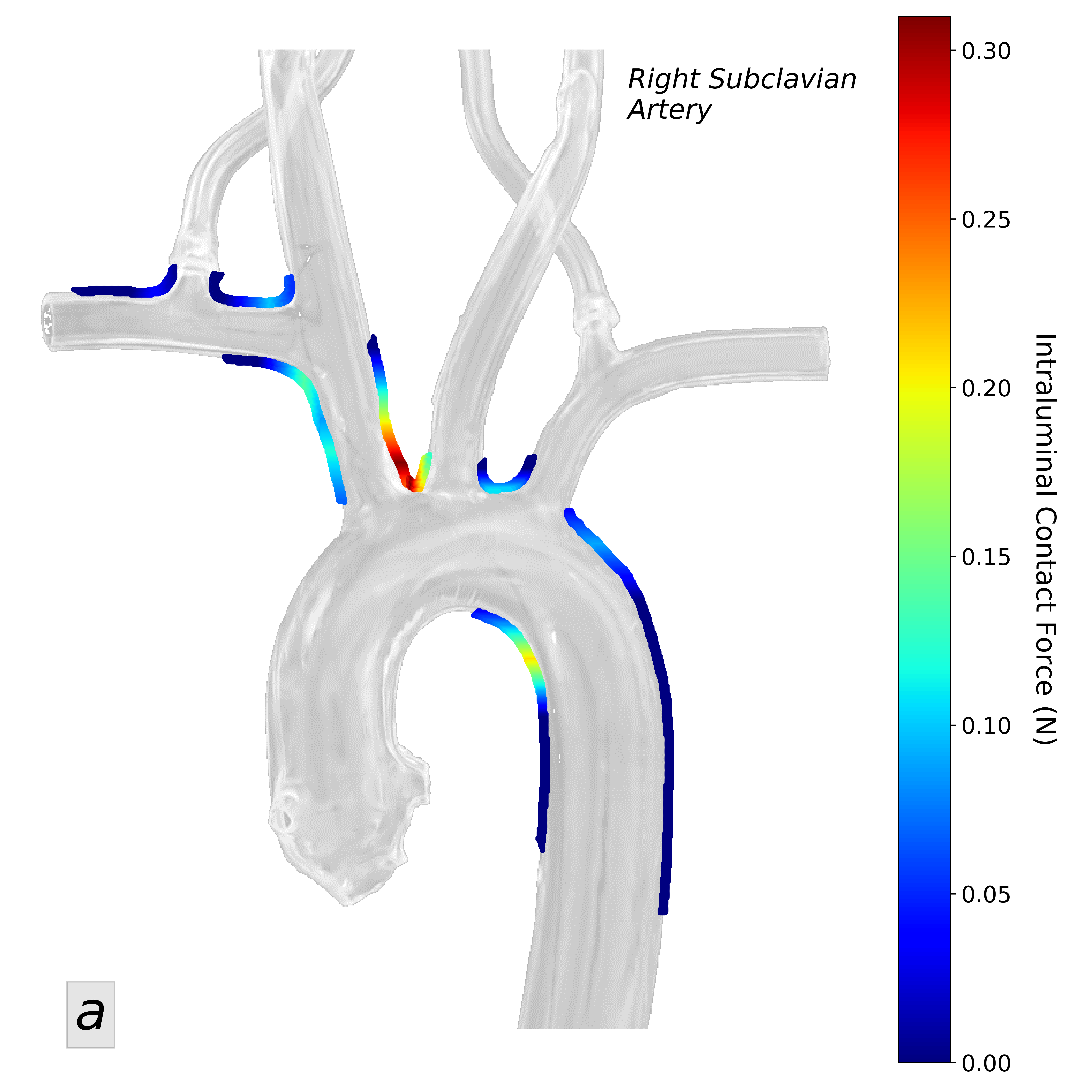}
	\includegraphics[width=0.32\textwidth]{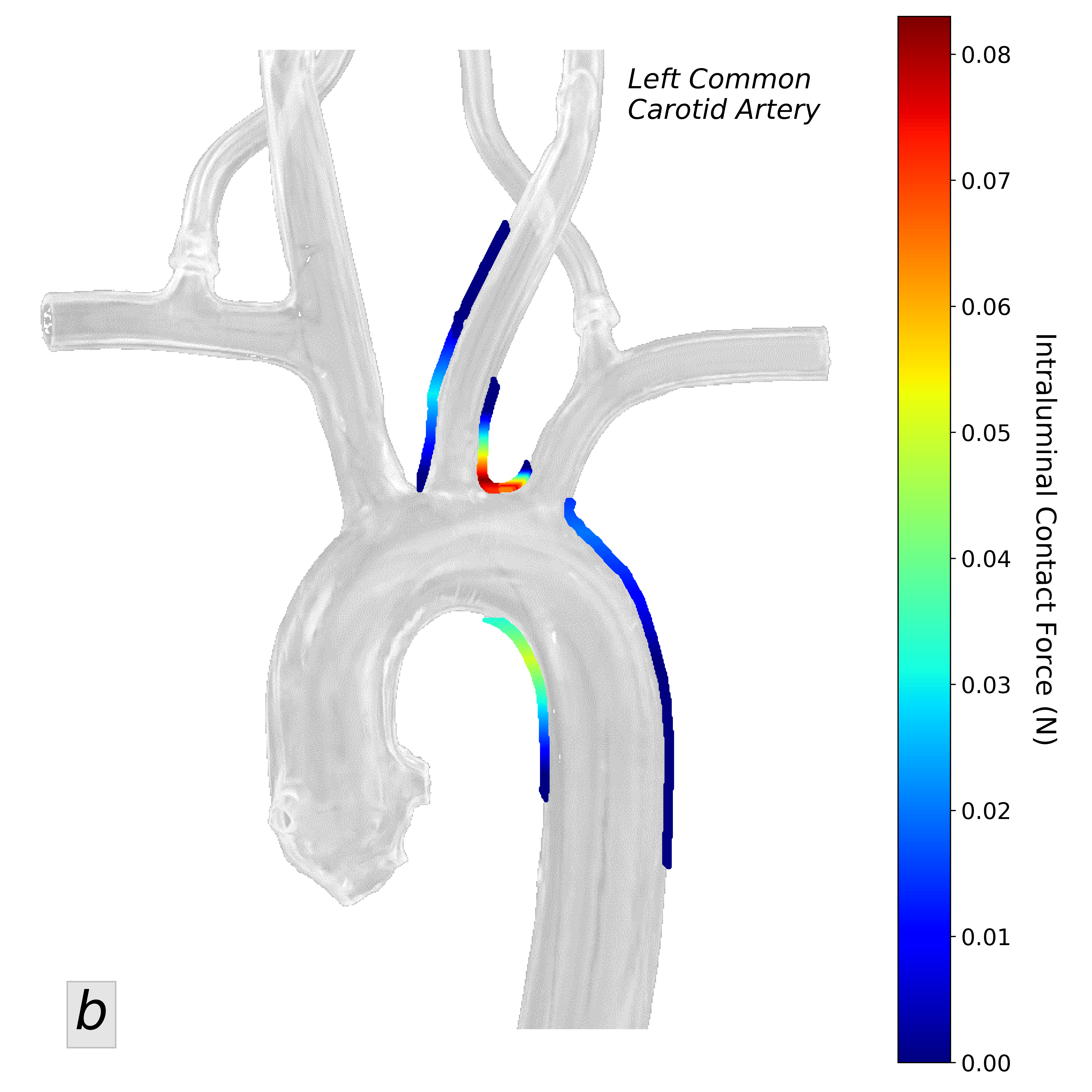}
	\includegraphics[width=0.32\textwidth]{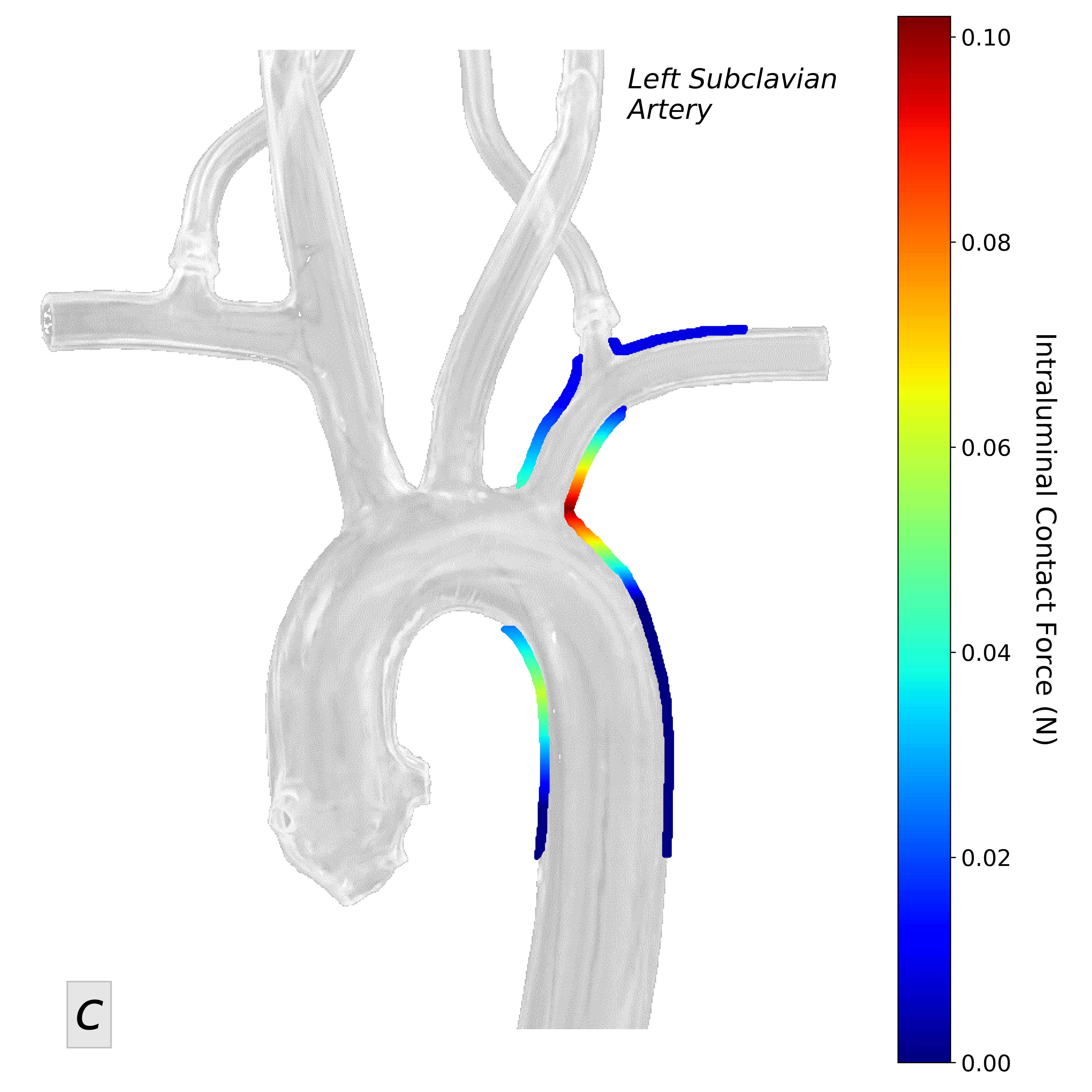}
	\captionsetup{margin=0.1cm}
	\caption{Intraluminal contact force monitoring results showing magnitude and position of applied forces on arterial wall through cannulation of Right Subclavian Artery (a), Left Common Carotid Artery (b), Left Subclavian Artery (c) (force scales are different in sub-figures).}
	\label{CFcountors}
\end{figure*}
A map of CF on vessel interaction boundaries over the time of the procedure is presented in Fig. \ref{RCCACountor} (results are for the same trial presented in Fig. \ref{CFsRCCA}). CF magnitudes are color-coded on their coordinate of interaction. The guidewire continuously interacts with the inner lumen arterial wall in which some parts of the wall get in contact several times. Maximum CF is visualized cases where several interactions occur on a part of the lumen. The path of interactions starts from both side of the descending aorta, gets to the aortic arch bifurcations and moves into RCCA where it mainly interacts with the right side of RCCA. Maximal contact forces are localized at the bifurcation edge of RCCA in the aortic arch, about 0.26\,N, and downward in the right side of the descending aorta, about 0.2\,N. This is due to the large deflection induced on the GW at those regions and also the guidewire's high rigidity modulus. 

Fig. \ref{CFcountors} depicts CF contour map of RSA, LCCA, and LSA cannulation. In these examples maximum CFs are 0.31N, 0.08N and 0.11N for RSA, LCCA and LSA, respectively. Observing the CF contours suggest that maximal CF in the procedures mainly happens at the bifurcation edges, where the guidewire path changes, and it is being forced to bend by the edge wall while advancing. In other words, the bifurcation edge wall acts as a support for the GW. Additionally, guidewire sticking to the phantom wall leads to more friction forces and consequently, more push force and CF. RSA contours looks similar to RCCA in terms of magnitudes and maximal CF locations on the vessels. This is because of common navigation procedure up to RCCA-RSA bifurcation, which consequently results in similar GW motion and interactions map. 

Table \ref{dataCF} shows intraluminal contact force metrics in all cannulation experiments. Cannulation of RCCA and RSA are associated with higher contact forces compared with LCCA and LSA, since these procedures are more dexterous and engage a longer portion of GW with larger deflections and larger friction forces. The highest CF peak up to 0.37\,N was observed in RSA cannulation. Average of peak CFs were 0.24\,N, 0.305\,N, 0.092\,N and 0.114\,N at RCCA, RSA, LCCA and LSA respectively and average of mean CF were 0.14\,N, 0.172\,N, 0.041\,N and 0.046\,N. The standard deviation of peak CF was from 0.021\,N to 0.053\,N showing a big variation over trials. It infers that not all procedures have small CF, but some may cause damage to arterial cells. Force time integration or force impact is a performance metric for quality of the procedure; a lower value indicates faster procedures with lower CF, i.e., the best desired scenario. Force impact was from 1.17\,N.s at LSA up to 9.85\,N.s at RSA. The higher value of CF and longer time of cannulation to reach the targeted point in RSA and RCCA resulted in significantly greater Force-Time integration\,(N.s).
\begin{table}[tb]
	\centering
	\captionsetup{margin=0cm}
	\caption{Mean values for computationally predicted intraluminal contact force and measured resultant force.}
	\def\arraystretch{1.2}
	\begin{tabular}{lllll}
		\hline
		& \multicolumn{1}{c}{RCCA} & \multicolumn{1}{c}{RSA} & \multicolumn{1}{c}{LCCA} & \multicolumn{1}{c}{LSA} \\ \hline
		Intraluminal Contact Forces    &                          &                         &                          &                         \\ \cline{1-1}
		Average Max CF (N)             & 0.24                     & 0.305                   & 0.092                    & 0.114                   \\
		STD Max CF (N)                 & 0.053                    & 0.046                   & 0.021                    & 0.026                   \\
		Average Mean CF (N)            & 0.142                    & 0.172                   & 0.041                    & 0.046                   \\
		Average F-T Integration (N.s)  & 8.57                     & 9.85                    & 1.23                     & 1.17                    \\ \hline
		Resultant Force                &                          &                         &                          &                         \\ \cline{1-1}
		Average Max force (N)          & 0.291                    & 0.321                   & 0.074                    & 0.081                   \\
		STD Max CF (N)                 & 0.082                    & 0.104                   & 0.034                    & 0.038                   \\
		Average Mean force (N)         & 0.078                    & 0.093                   & 0.031                    & 0.035                   \\
		Average F-T Integration  (N.s) & 4.75                     & 6.45                    & 1.06                     & 0.875                  
	\end{tabular}
	\label{dataCF}
\end{table}
\subsection{Estimation error}
\begin{table}[bt]
	\centering
	\caption{Model accuracy based on shape estimation error along with GW lengths and deflections.}
	\def\arraystretch{1.3}
	\begin{tabular}{llllll} 
		\hline
		\multicolumn{2}{l}{}                     & RCCA & RSA & LCCA & LSA  \\ 
		\hline
		& Max length\,(mm)                       & 314  & 332 & 257  & 193  \\
		& Mean length\,(mm)                      & 211  & 224 & 176  & 124  \\
		& Mean deflection\,(mm)                  & 63   & 75  & 51   & 42  \\
		\multicolumn{2}{l}{FEM Estimation Error ($ E_p $)}   &      &     &      &      \\
		&   $RMSE_{u}\,(mm)$                             & 2.2  & 3.1 & 1.8  & 1.3  \\
		&  $ MAXE_{u}\,(mm) $                              & 4.3  & 5.7 & 3.2  & 2.4 
	\end{tabular}
	\label{FEMerror}
\end{table}
A validation of the proposed method showing accuracy and effectiveness in multi-point force estimations were performed in a earlier study \cite{razban2018sensor}. It suggested that a smaller error in force estimation was associated with a smaller shape simulation error. In this study, direct intravascular force measurement is not possible, thus the performance evaluation is conducted based on shape estimation error by comparing actual GW deflected shape and FEM simulation. Table \ref{FEMerror} includes maximum and average root-mean-square-errors in shape estimation along with GW lengths values and average applied deflections in contact points. The estimation error is defined in Eq. \ref{Error}. Within all arteries, maximal errors are 4.3\,mm and 5.7\,mm for the RCCA and RSA cases, respectively, where the maximum GW lengths are 314\,mm and 332\,mm. Average applied deflections are between 42\,mm and 75\,mm within the arteries. The estimation error is higher for scenarios with larger deflection.
\subsection{Resultant exerted forces}
\begin{figure}[] 
\centering
\begin{subfigure}{0.25\textwidth}
\includegraphics[width=\textwidth,trim={0.5cm 0cm 1.3cm 1cm},clip]{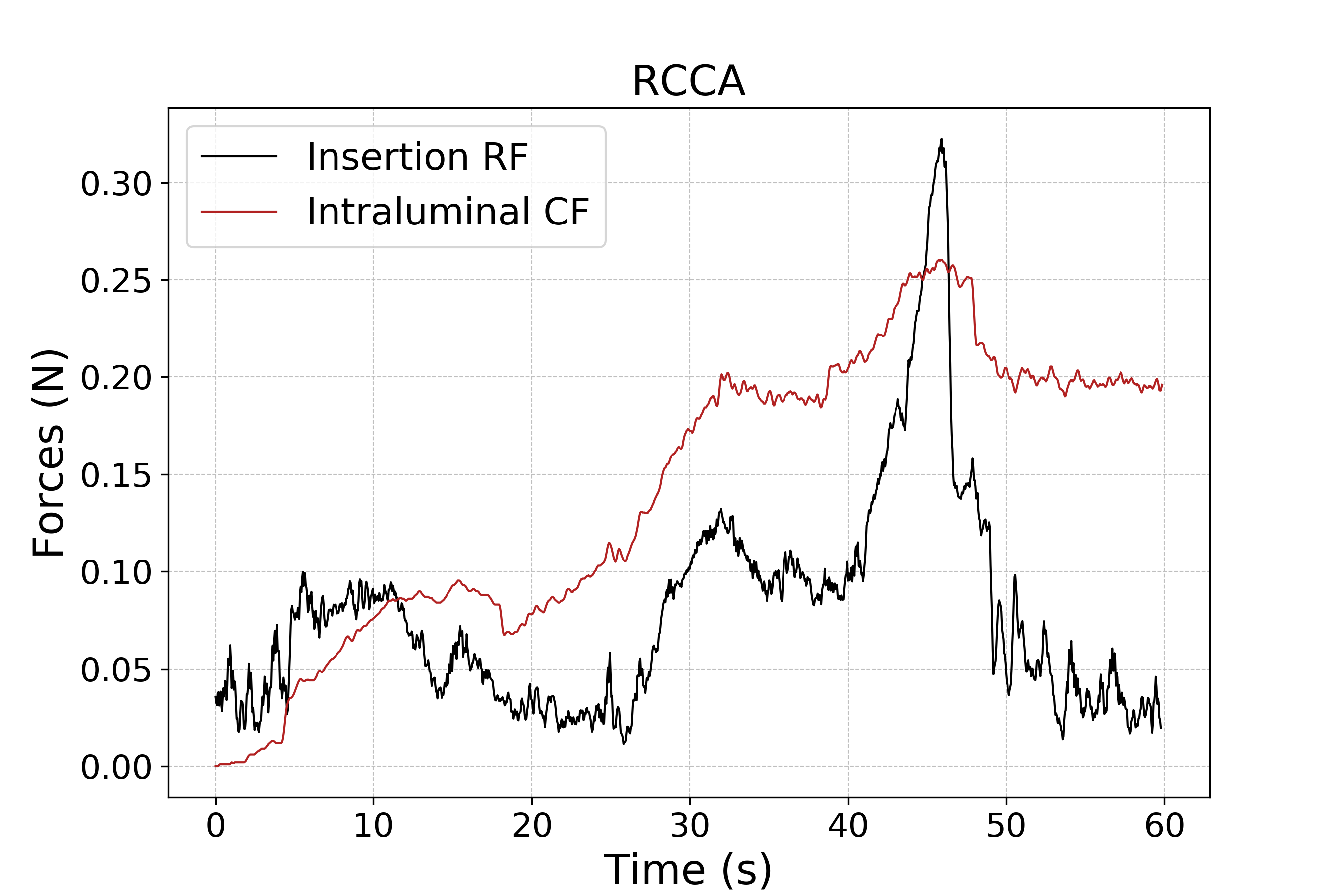}
\end{subfigure}%
\begin{subfigure}{0.25\textwidth}
\includegraphics[width=\textwidth,trim={0.5cm 0cm 1.3cm 1cm},clip]{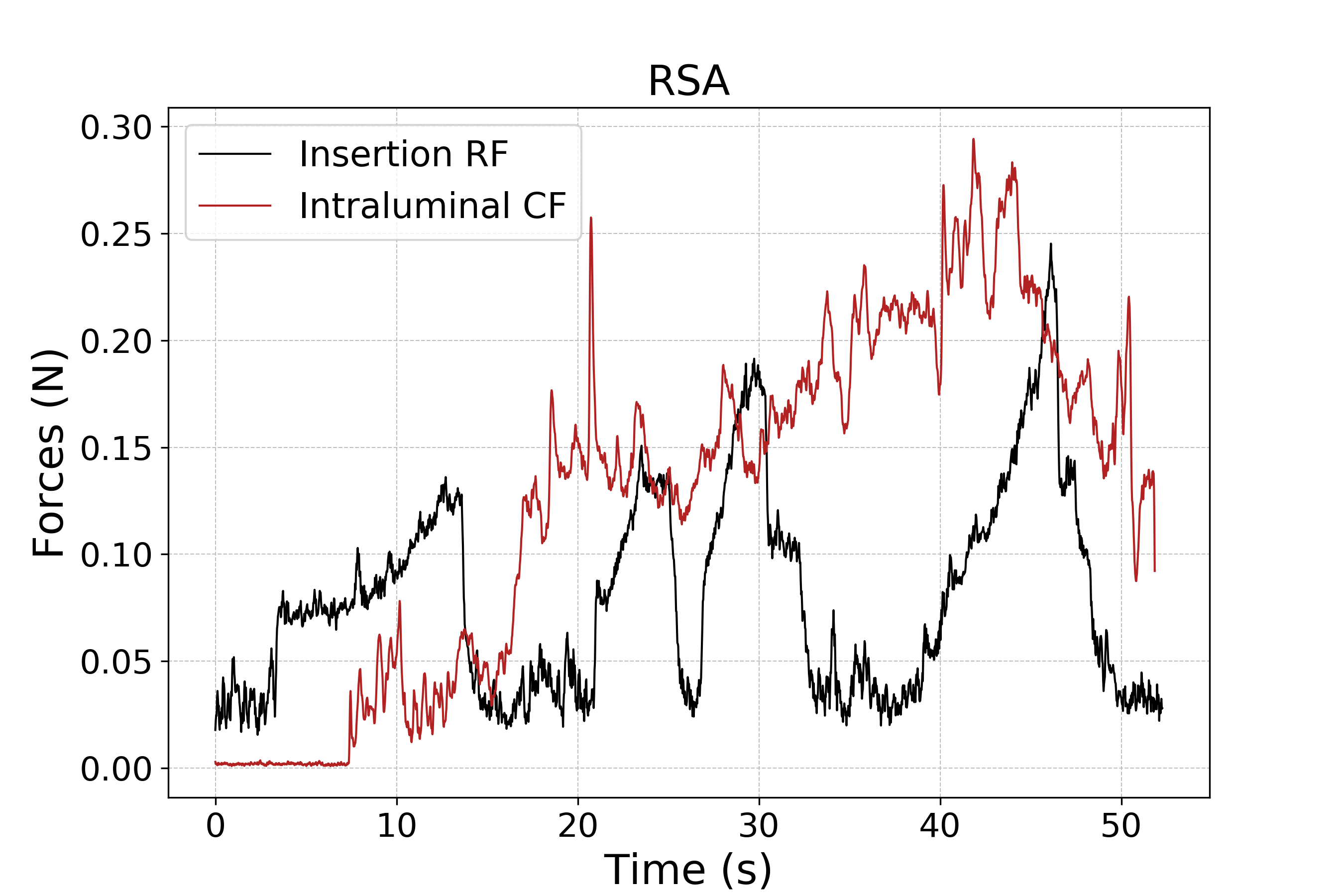}
\end{subfigure}%
\vspace{0.1cm}
\begin{subfigure}{0.25\textwidth}
\includegraphics[width=\textwidth,trim={0.5cm 0cm 1.3cm 1cm},clip]{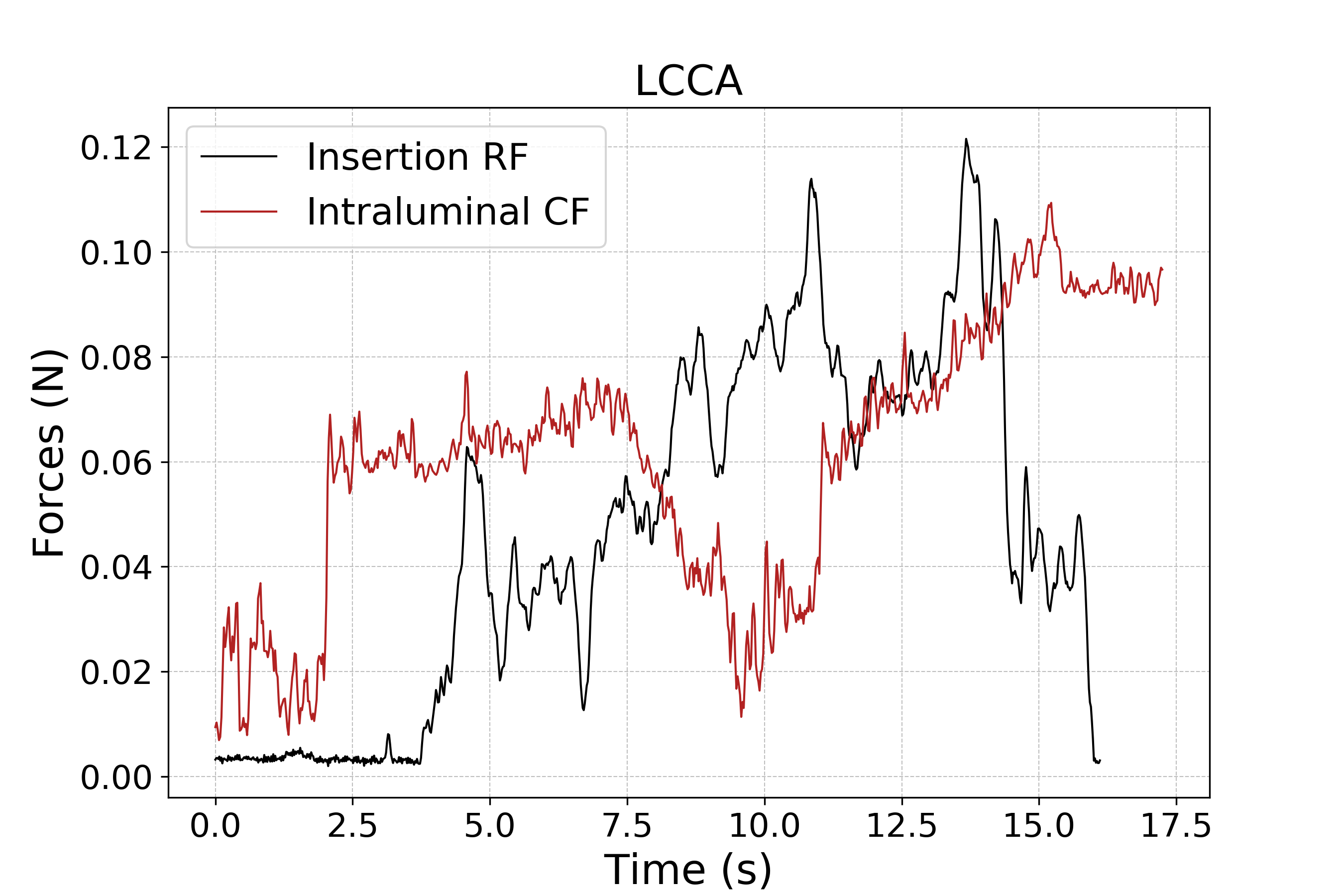}
\end{subfigure}%
\begin{subfigure}{0.25\textwidth}
\includegraphics[width=\textwidth,trim={0.5cm 0cm 1.3cm 1cm},clip]{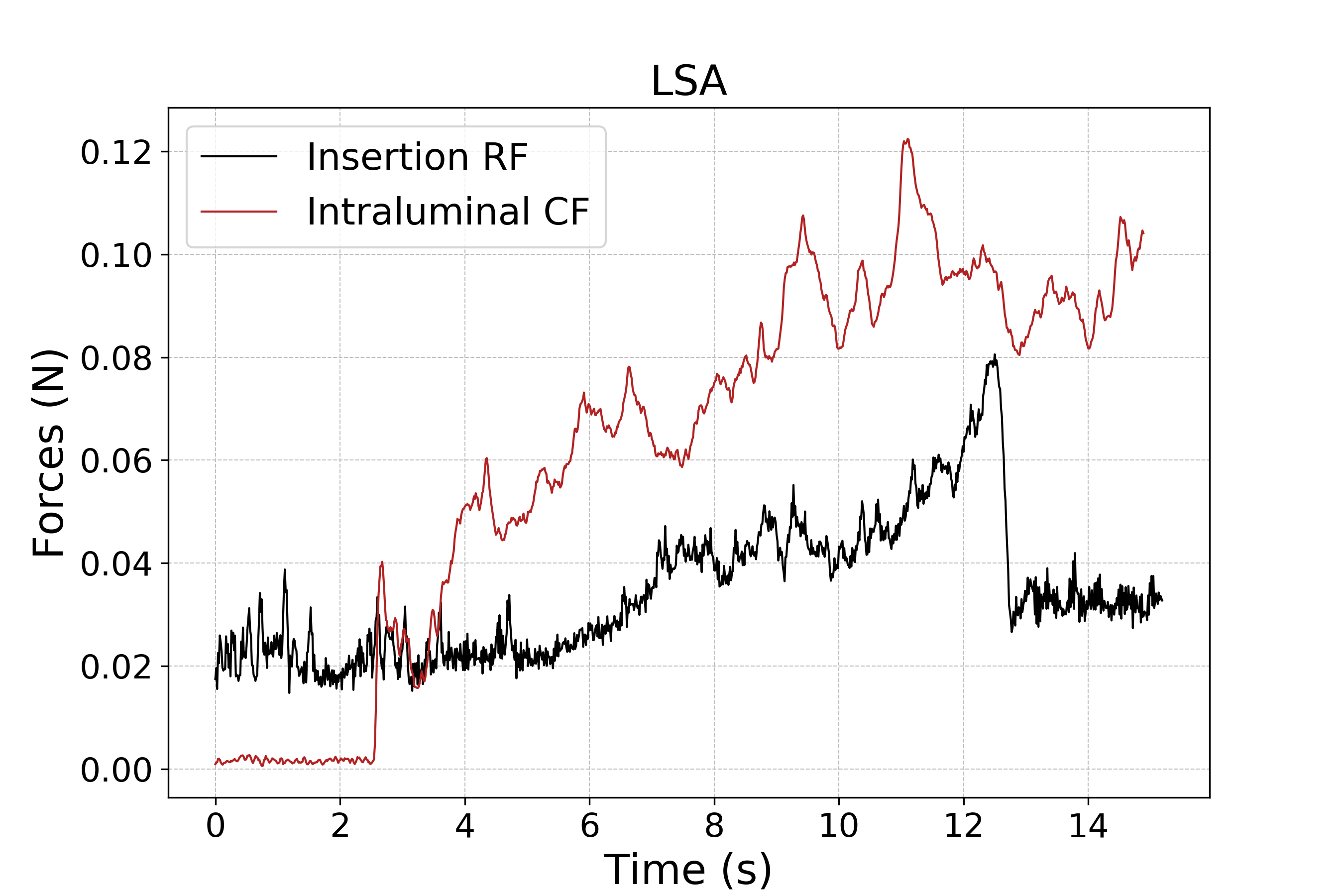}
\end{subfigure}%
\captionsetup{margin=0cm}
\caption{Total force exerted on the phantom, RF, is compared with maximum local intraluminal CF in cannulation of arteries.}
\label{CFvsSensor}
\end{figure}

Some studies proposed the measurement of total catheter/GW insertion forces or total contact forces applied on a phantom for either contact force evaluation or insertion force control \cite{jayender2008autonomous, rafii2016reducing}. Total catheter insertion force does not represent the local CF applied on the inner arterial wall, while it is likely to be the resultant of intraluminal CFs plus unknown friction forces from introducer sheet, access point, etc. Here, we are comparing maximal intraluminal CF with resultant exerted force (RF) on the phantom recorded by the F/T sensor platform to highlight the potential need for intraluminal interaction monitoring. Fig. \ref{CFvsSensor} shows samples of force comparisons within arteries. There is no consistent trend; the maximal intraluminal CF is higher or lower than RF; however, in the majority of times, intraluminal CF is larger. RF may fall through procedure while CF remains large. It also suggests that RF is not a promising indication of complications since it is not as high as the CF, and their peaks do not match. Table \ref{dataCF} includes intraluminal CF and RF metrics. Max RF values may be higher compared to max CF, whereas mean RF is always showing lower values. This means that low catheter/GW insertion force could increase deflection gradually, which may lead to a high intraluminal CF with a risk of arterial wall injury. Monitoring and controlling RF, or insertion force, might smoothen the procedure, but it does not necessarily prevent excessive CF and subsequent complications.
\subsection{Stress}
\begin{figure}[t]
	\centering
	\includegraphics[width=0.47\textwidth,,trim={0cm 0cm 7.2cm 0cm},clip]{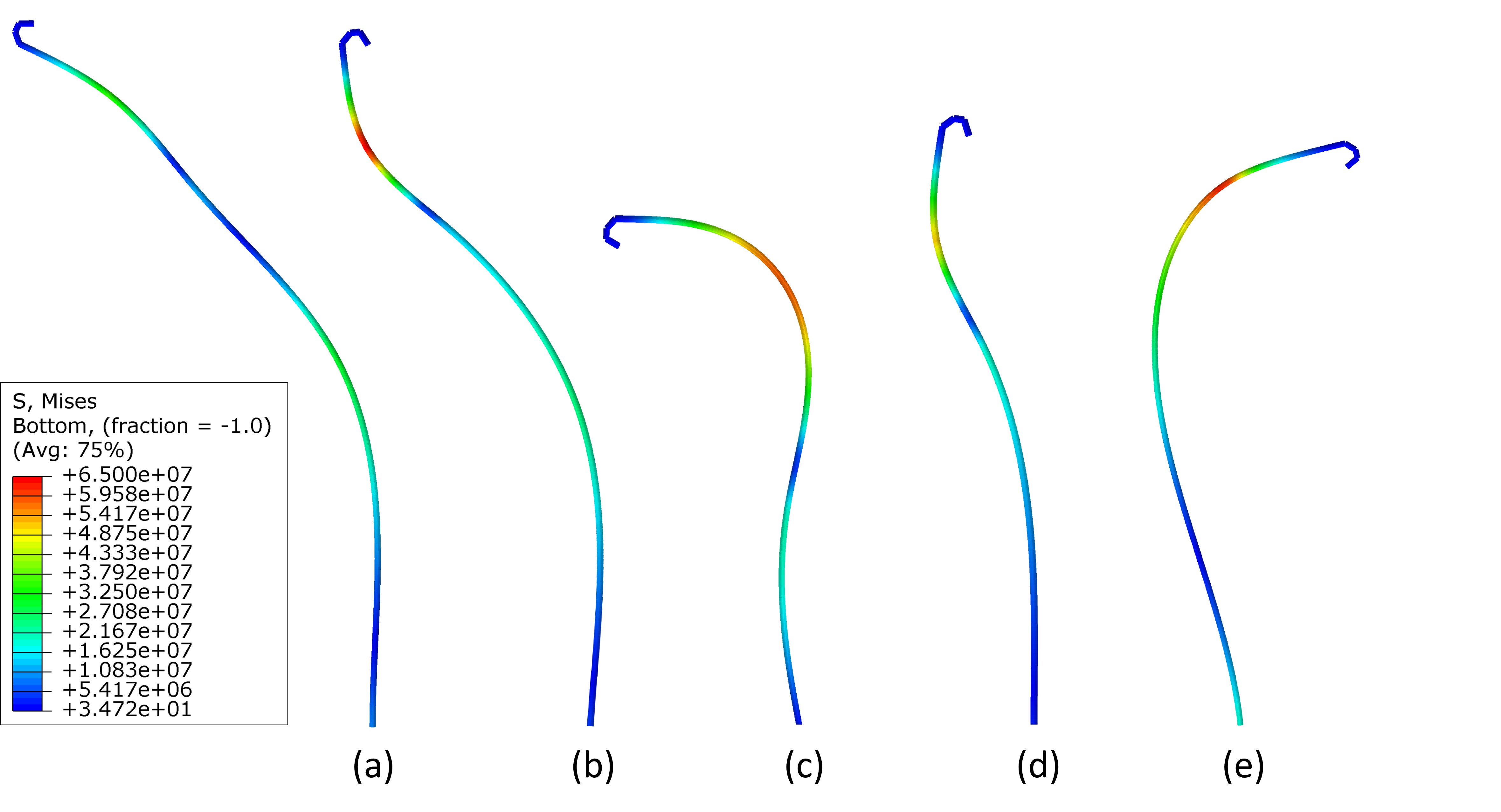}
	\caption{Stress estimation of GW during cannulation of arteries, RSA (a), RCCA (b), aortic arch (c), LCCA (d), LSA (e).}
	\label{StressFEM}
\end{figure}

In the proposed method, we can extract stress and strain from the GW/catheter FEM model at any instant of time. Fig. \ref{StressFEM} presents samples of GW stress contour in arteries. Using intraluminal information, GW structure can be optimized to maximize navigation performance while minimizing CF and stress.
\subsection{Discussion and limitations} 
The observed errors in estimation could be due to measurement error in GW bending modulus over its length. Sharp changes in flexural rigidity may not be appropriately seen in sequential experiments. However, manufacturers have bending information based on their design, which can be used if such information is provided. Another error source is that the GW model is initially considered as a straight cantilever, whereas small deflection of the structure is observed for high initial lengths. The accuracy of force monitoring varies depending on the correctness of image segmentation as well, but is not sensitive to it. Camera calibration could cause such errors in this study, which may not be an issue in commercially available clinical imaging systems.
The proposed force estimation does not require information on arterial wall material properties and tissue deflections; however, having that information would help to estimate local pressure profile on contact. It is almost certain that a distributed contact may happen but it should result in a wider contact area, which produces lower pressure and risk. In this study, the amount of force, deflection, stress, and other parameters are reported for navigation of a GW in a training phantom, therefore, they are not necessarily as the same magnitude as in clinical practice, and such conclusion should not be made. 
\section{Conclusion}
Tool-vasculature interaction was reported to be the leading cause of complications and major concern, such as embolization, stroke, ischaemic lesion, and perforation, in endovascular interventions. In this study, the case of catheter contact point forces is of interest, as it is the extreme condition with a high risk of over-pressure on the vasculature. We implemented an image-based intraluminal catheter-vessel interaction monitoring tool based on imaging data and numerical computation. The image segmentation algorithms successfully detect and track contacts, vasculature boundaries, catheter and compute needed pose measurements. The FEM model effectively simulates manipulation and predicts contact forces and structural stress. Remote cannulation of the aortic arteries was performed using a robotic unit and intraluminal contact force monitoring was achieved by tracking local CFs and building a contour map of CF on the arterial wall. Results suggest that RCCA and RSA associated with higher CF where maximal CF happens at the bifurcation edge of the aortic arch. The estimation error was low, showing the fidelity of the model. Contact forces can be visualized intraoperatively for clinicians to prevent injuries and reduce the learning curve for novices. CF could be used by the robotic system as a feedback for restrained force control. Resultant forces exerted on the vascular phantom is directly measured and compared with local CF trend, where small RF and catheter insertion forces could be detected while ICF was large on the vessel wall. The model can predict structural stress of the GW besides CF data in practice, which can help to research and optimize the design of interventional tools. Nearly all cardiovascular procedures are under 2D real-time imaging. Physicians keep the imaging plane tangent to catheter motion, which we have simulated experimentally in this study. Extending the proposed methodology to 3D is an intuitive step that requires an upgrade to a 3D imaging platform and 3D FEM modeling. The proposed method can be further implemented on other types of interventional devices and cardiovascular procedures.
\section*{Acknowledgment}
Authors would like to thank the Surgical Innovation Program and McGill University Health Centre for clinical visits.  
\addtolength{\textheight}{0cm} 
\ifCLASSOPTIONcaptionsoff
  \newpage
\fi


%
\bibliography{REF}
\bibliographystyle{IEEEtran}
%

%
%
%




\end{document}